\documentclass[a4paper,fleqn]{cas-sc}
\usepackage[authoryear]{natbib}

\usepackage{import}

\usepackage{tabularx}
\usepackage{graphicx}
\usepackage{float}
\usepackage{enumitem}
\usepackage{pdfpages}
\usepackage{placeins}
\usepackage[font=it,skip=3pt]{subcaption}
\usepackage{threeparttable}

\makeatletter
\g@addto@macro\UrlBreaks{\do\-}
\makeatother

\newcommand{\bm}[1]{$\boldsymbol{#1}$}
\newcommand{\bt}[1]{\textbf{#1}}
\renewcommand{\it}[1]{\textit{#1}}
\newcommand{\mt}[1]{\texttt{\detokenize{#1}}}

\graphicspath{{.}}

\begin{document}

\let\WriteBookmarks\relax
\def\floatpagepagefraction{1}
\def\textpagefraction{.001}

\shorttitle{Spatiotemporal Multi-Task Graph Transformer for Trip-Level Transit Prediction}
\shortauthors{O. Yusuf et~al.}
\title [mode = title]{Spatiotemporal Multi-Task Graph Transformer for Trip-Level Transit Prediction}

\author[1]{Oluwaleke Yusuf}[orcid=0000-0002-5904-648X]
\author[1]{Adil Rasheed}[orcid=0000-0003-2690-983X]
\author[2]{Frank Lindseth}[orcid=0000-0002-4979-9218]

\affiliation[1]{
    organization={Department of Engineering Cybernetics, Norwegian University of Science and Technology (NTNU)},
    city={Trondheim},
    citysep={},
    postcode={NO-7491},
    country={Norway}
}
\affiliation[2]{
    organization={Department of Computer Science, Norwegian University of Science and Technology (NTNU)},
    city={Trondheim},
    citysep={},
    postcode={NO-7491},
    country={Norway}
}

\begin{abstract}[S U M M A R Y]
    Passenger count data from public transit systems reveals urban mobility patterns and is essential for planning, operation, and optimisation. However, non-linear spatiotemporal interdependencies across stops and lines make modelling and prediction challenging. Existing approaches often rely on fixed temporal, spatial, or stop-level formulations, limiting their ability to capture within-trip evolution and network context.
    This study proposes SMT-GraphFormer, a spatiotemporal multi-task graph transformer that frames trip-level transit prediction as sequence-to-sequence modelling. Given a line's stop sequence and trip-level context, the model predicts successive boarding and alighting counts, with delay and dwell time treated as encoder-side surrogate tasks. Key components include graph embeddings for multi-relational stop similarity, a context encoder for weather and temporal information, and a multi-gate mixture-of-experts module that produces task-specific decoder representations for boarding and alighting predictions.
    Evaluation on public bus transit data from Trondheim, Norway, shows that SMT-GraphFormer outperforms stop-level tabular benchmarks, with ablation studies examining each component's contribution. The sequential formulation yields substantial gains on alighting prediction ($+$0.24 in $R^2$) and consistent improvements on boarding, delay, and dwell, confirming the value of explicit trip-level sequential bias and inter-target dependencies.
    These findings demonstrate the potential of transformer-based sequence modelling for capturing complex spatiotemporal dynamics in public transit and underscore the value of architectures tailored to transit data rather than off-the-shelf tabular models. The proposed framework provides a horizon-agnostic basis for scenario analysis in digital twin environments, supporting informed decision-making by planners and transit operators.
\end{abstract}

\begin{keywords}
    Public Transit \sep
    Demand Prediction \sep
    Transformer Networks \sep
    Deep Learning \sep
    Spatio-temporal Graphs \sep
    Sequence Modelling
\end{keywords}

\maketitle

\section{Introduction}
Population growth in urban areas has placed increasing strain on mobility systems and their capability to serve residents adequately. With urban populations expected to continue growing rapidly \citep{Gulc2024cos}, the sustainability of contemporary mobility solutions and their impact on urban quality of life have become central concerns \citep{Ritchie2020cpt}. Mobility policy is increasingly shifting away from car-centred approaches towards people-centred planning \citep{Millard_Ball2011awr,Jones2016teo}, prioritising public transit and active mobility over private car use. This shift underscores the need to improve public transit services and make integration with active modes more seamless, helping residents choose more sustainable forms of mobility.

These trends have also intensified the need for data-driven historical analysis and predictive modelling of passenger counts and operational metrics in urban public transit systems \citep{Egu2021mtp}. Detailed analyses and accurate predictions of passenger counts and scheduling deviations at the stop and trip level support urban planners allocating resources, transit operators adjusting schedules, and travellers planning journeys. Focusing on public bus transit, this study addresses the prediction of boarding and alighting passenger counts at each stop along a trip, together with delays relative to scheduled arrival times and dwell times. These quantities are critical for understanding demand patterns, operational performance, and passenger experience, and are interdependent through complex spatiotemporal dynamics.

There have been various approaches to modelling these quantities across short- and long-term horizons in urban public transit systems such as bus \citep{Kong2024ebs,Shrivastava2024dlb}, metro \citep{Hu2024gte}, and scooter \citep{Xu2023rtf} networks. However, such approaches face difficulties with the complex non-linearities and dynamic spatiotemporal interdependencies of human mobility flows. This is further complicated by the interplay of various factors intrinsic and extrinsic to the transit system.
Extrinsic influences such as weather conditions, urban characteristics, natural terrain, and temporal information influence mobility patterns \citep{Wei2022iti,Verma2021esc,Yang2023aio}. Intrinsically, the target variables are coupled through cascading dependencies: dwell time at a stop rises with the number of boarding passengers; the resulting delay propagates to the next stop; alighting at that stop depends on earlier boarding; and higher alighting in turn extends dwell. These chains imply that predictions at successive stops cannot be treated independently but must be modelled as an ordered sequence within a trip. Furthermore, there are interdependencies across lines at the network level, where shared stops and transfer points create complex spatial relationships that influence passenger counts.

Previous work on predictive modelling of passenger counts and operational metrics \citep{Yusuf2025ddp} has shown that stop-level modelling alone lacks inductive biases on the spatial structure of transit networks and cannot adequately capture relationships between boarding and alighting counts without explicit sequential context.
In addition, studies on transit-related modelling typically focus on fixed spatial or temporal aggregation levels, such as origin--destination (OD) pairs \citep{Li2022gnn,Hu2024gte}, predefined time intervals \citep{Shrivastava2024dlb,Kong2024ebs}, or spatial regions \citep{Xu2023rtf,Li2022gnn}. This aggregation overlooks the evolution of passenger counts at the stop and trip level, disregarding spatial interdependencies between successive stops and their local attributes. Temporal aggregation reduces the temporal resolution of predictions, limiting insights into the effects of operational metrics such as delays relative to scheduled arrival and dwell times at stops. Treating passenger counts solely as a time-series problem limits the model's predictive window and generalisation over time, reducing its applicability for \it{what-if?} scenario analyses.

To address these limitations, this study reframes the trip-level prediction of passenger counts and operational metrics as a sequence modelling problem akin to natural language modelling. Individual trips are treated as sentences, individual stops as tokens, and the prediction of boarding and alighting counts proceeds autoregressively from the first stop to the last. The proposed framework, \mt{SMT-GraphFormer}, uses a modified encoder-decoder transformer architecture whose input is a comprehensive representation of a trip, including the line and stop sequence, stop-level features such as natural terrain and urban characteristics, and trip-level context such as weather conditions and temporal information. The encoder processes this input to produce contextual stop representations and predicts delay and dwell times as surrogate tasks. The decoder then produces autoregressive states for boarding and alighting prediction, with a multi-gate mixture-of-experts module producing task-specific decoder representations that feed into the prediction heads.

This formulation allows the model to learn the complex spatiotemporal dynamics of passenger counts across stops within a trip, while also integrating operational metrics and external factors without being constrained to fixed temporal or spatial aggregation levels. This horizon-agnostic approach is well-suited for scenario analysis in digital twin environments, where urban planners and transport operators can vary input parameters such as trip schedules, stop sequences, urban characteristics, and weather conditions, then simulate the resulting evolution of passenger counts and operational metrics across the trip. Thus, the transformation of stop-level tabular data into trip-level sequential format, along with feature engineering at the stop, trip, and network levels, is a critical component of the proposed approach. The main contributions of this study are as follows:

\begin{enumerate}[itemsep=0pt]
    \item \bt{Data Pipeline:} Preprocessing pipeline that transforms stop-level tabular transit records into trip-level sequential representations. Terrain, urban, and weather features are engineered from external sources at the stop and trip level, alongside relational stop-similarity matrices for the graph autoencoder.
    \item \bt{Model Architecture:} Modified encoder-decoder transformer predicting boarding, alighting, delay, and dwell jointly across a trip sequence. Key \mt{SMT-GraphFormer} components include spatiotemporal graph embeddings, a trip-level context encoder, encoder-side surrogate-task supervision, and a multi-gate mixture-of-experts module for task-specific decoder representations.
    \item \bt{Empirical Validation:} Evaluation on one month of real-world data from the public bus transit system in Trondheim, Norway. Results demonstrate that the sequential formulation outperforms stop-level tabular benchmarks on all four targets, supported by ablation studies examining each component's contribution.
    \item \bt{Generalised Framework:} Horizon-agnostic predictive framework that, given a line's stop sequence and trip-level context, generates the full per-stop trajectory of passenger counts and operational metrics. Enables \it{what-if?} scenario analysis in digital twin environments without reliance on fixed temporal windows.
\end{enumerate}

The rest of this paper is organised as follows:
\autoref{sec:related} reviews the relevant literature on transit prediction, use of graph-based embeddings and transformer architectures, and integration of external factors.
\autoref{sec:m-data} details the data preprocessing pipeline for transforming raw stop-level records into the trip-level sequential format, along with the feature engineering of stop-level attributes, trip-level context, and relational matrices for graph embeddings.
\autoref{sec:m-model} describes the \mt{SMT-GraphFormer} architecture, model training and inference protocols, and the benchmark models used for comparison.
\autoref{sec:results} presents the experimental results and comparison with benchmarks, in addition to the ablation studies on model components and the analysis of autoregressive inference.
\autoref{sec:conclusion} summarises the key findings, discusses the implications and limitations of the results, and outlines potential directions for future research.

\section{Related Work}
\label{sec:related}
This section reviews the relevant literature across four themes that inform the design of \mt{SMT-GraphFormer}: \emph{(i)} predictive modelling of mobility-related variables, \emph{(ii)} graph-based embeddings for encoding spatiotemporal relationships between transit system elements, \emph{(iii)} transformer architectures that integrate graph representations, and \emph{(iv)} the role of external factors in shaping mobility patterns and thus improving predictive performance.

\subsection{Mobility-Related Predictive Modelling}
There have been numerous studies on predicting passenger counts across short- and long-term horizons in urban public transit systems such as bus \citep{Kong2024ebs,Shrivastava2024dlb,Li2022gnn,Kong2022ehm}, metro \citep{Hu2024gte}, scooter \citep{Xu2023rtf}, and bike-sharing \citep{Ruhmann2024ibs} networks.
There have also been similar studies in mobility-adjacent domains dealing with traffic flow \citep{Zhang2024tta,Zhao2024gst,Dong2024mms,Zeng2023mdt}, bus arrival time \citep{Jeong2024btp,Alam2021pii}, demand distribution \citep{Guo2023stc}, daily human mobility \citep{Wang2024ldh,Corrias2023eta,Sun2022tna}, origin--destination (OD) estimates \citep{Li2023ccq}, road correlations \citep{Dong2022llr}, and transit mobility structure \citep{Zhang2021utm}, amongst others.
However, such studies are often limited in their ability to capture the complex spatiotemporal dependencies and non-linearities inherent in human mobility dynamics, which are challenging to analyse, model, and predict accurately and consistently.

This challenge has led to the development of various methodological approaches with varying degrees of efficacy. The demonstrated ability of transformer architectures, first introduced by \citet{Vaswani2017aia}, to model long-range dependencies in sequential data has increasingly been leveraged in the transit domain.
Within bus transit prediction specifically, \citet{Li2022gnn} and \citet{Hu2024gte} aggregate passenger counts across origin--destination pairs, modelling demand between pairs of stations or zones. \citet{Shrivastava2024dlb} cluster transit segment relations based on ridership patterns and train separate LSTM models per cluster to predict occupancy across time bins. \citet{Kong2022ehm} and \citet{Kong2024ebs} construct stop relationship networks and apply graph-based deep clustering to extract mobility patterns for stop-level count prediction. \citet{Pei2023bpf} decompose bus passenger count time series using wavelet packets and feed the components into a bidirectional LSTM with an attention mechanism.

Limited data granularity and the complexity of the underlying dynamics push these approaches towards a common compromise: prediction is framed at an aggregated level, whether spatially across OD pairs, temporally across predefined intervals, or independently at individual stops. None of them explicitly model the sequential evolution of passenger counts across the ordered stops of a single trip, where boarding and alighting at each stop are conditioned on the preceding sequence. This trip-level sequential structure is central to the formulation proposed in our study.

\subsection{Graph Embeddings for Transit Systems}
When applying transformer architectures to transit prediction, a key consideration is how to encode the various elements of the transit system in a way that captures their spatiotemporal relationships and provides rich information on stop-, line-, and network-level mobility dynamics. While standard learnable embedding layers \citep{Dong2024mms,Abideen2021dws,Hu2023hdp} and field embeddings \citep{Guo2023stc} can be incorporated into the transformer architecture, these approaches cannot accurately capture the true spatiotemporal dependencies of mobility flows, as noted by \citet{Zhang2024tta}. This has motivated the integration of graph-based embeddings aimed at exploiting the inherent graph structure of transit systems.

Such embeddings are derived from relational matrices that encode pairwise relationships and correlations between transit system elements at various levels of granularity. Beyond basic binary adjacency matrices, more informative variants include distance-weighted matrices based on threshold Gaussian kernels \citep{Zhao2024gst,Kong2024ebs}, demand similarity matrices derived from Pearson correlations of historical passenger demand sequences \citep{Li2022gnn,Shrivastava2024dlb}, matrices based on dynamic time warping of traffic volume time series \citep{Zeng2023mdt}, attribute similarity matrices encoding local static attributes of nodes \citep{Zhang2021utm,Xu2023rtf}, and cosine similarity matrices derived from traffic volume characteristics \citep{Dong2022llr}. Several studies construct multiple such matrices to capture different relational aspects simultaneously \citep{Xu2023rtf,Zhao2024gst,Zeng2023mdt}.

To convert these relational matrices into dense, low-dimensional embeddings, various strategies have been employed. Random walk algorithms such as DeepWalk and Word2Vec treat random walks on the graph as sentences for language model training \citep{Sun2022tna,Guo2023stc}. \citet{Zhang2021utm} proposed a joint graph embedding framework using autoencoders to learn shared low-dimensional representations from both attribute similarity and transit mobility pattern matrices.
In addition, autoencoders based on graph convolutional networks (GCNs) have been used to cluster mobility patterns from passenger networks based on shared stop usage \citep{Kong2022ehm} and stop networks based on inter-stop distance \citep{Kong2024ebs}. \citet{Dong2022llr} used trajectory next-hop prediction models to learn dynamic road embeddings that capture correlations beyond direct route connectivity. Three-dimensional convolutions have also been applied to jointly extract spatial and temporal features from structured traffic data \citep{Zhang2024tta}.

Our framework builds on this line of work by constructing multiple relational matrices to capture stop- and line-level relationships, including distance-based, demand-based, and attribute-based correlations. These matrices are then encoded through a graph autoencoder into dense embeddings that capture the spatial structure and temporal dynamics of the transit network.

\subsection{Graph Transformer Architectures}
The integration of graph-based embeddings into transformer architectures has been explored in various studies, motivated by the complementary strengths of graph representations for spatial relationships and transformers for sequential modelling.
\citet{Zhao2024gst} proposed a traffic flow prediction framework that represents sensor data as nodes in a graph, extracting spatial features with a multi-view graph convolutional network (GCN) that constructs adjacency matrices reflecting human mobility patterns. The temporal component utilises a transformer, along with Convolutional LSTM (ConvLSTM) to enhance temporal perception.
\citet{Xu2023rtf} applied this approach to scooter-sharing demand forecasting, where the GCN module integrates four graph types to model spatial adjacency, functional similarity, demographic similarity, and transportation supply similarity between urban zones.
In \citet{Wang2024ldh}, a self-attention embedding mechanism was used to map regional attributes into location embeddings, enabling a transformer to learn and generate the daily mobility sequences of individuals.

\citet{Hu2024gte} integrated complex network indicators to dynamically capture spatial dependencies between subway stations within a transformer. The multi-gate Mixture-of-Experts (MMoE) module was used to model the interdependencies between entry and exit passenger counts for improved predictive performance.
In \citet{Zhang2024tta}, the traditional fixed positional encodings of the transformer were replaced with learnable encodings, allowing the model to adaptively fit different temporal trends and horizons. A spatiotemporal embedding mechanism was also introduced for integrating information from external factors.
\citet{Guo2023stc} addressed demand forecasting during the COVID-19 pandemic with a transformer encoder that models high-order feature interactions between demand distributions and spatiotemporal features, incorporating a population mobility graph for spatial dependencies and an auxiliary pandemic trend prediction task.

Our architecture also combines graph-based embeddings with a transformer, but differs from these studies principally in its use of the encoder-decoder formulation. Where the studies above use graph-enhanced transformers for predictions at fixed aggregation levels, our architecture uses the encoder to build contextual representations and the decoder for autoregressive predictions across the ordered stops of a trip. This formulation allows us to explicitly model the sequential evolution of passenger counts across a trip, where boarding and alighting at each stop are conditioned on the preceding sequence, which is not captured by stop-level or aggregated approaches.

\subsection{Integrating External Factors}
The spatiotemporal dynamics of human mobility are influenced by a wide range of factors beyond the mobility system itself. Externalities such as weather conditions \citep{Wei2022iti,Tian2024qwi,Ngo2024tio}, residential densification \citep{Kim2021bmh}, urban structures \citep{Verma2021esc}, the built environment \citep{Yang2023aio,Zhang2025tnl,Liu2023tit}, and the COVID-19 pandemic \citep{Zeb2024fpt} have all been shown to influence mobility patterns.
Thus, several of the aforementioned studies -- \citet{Wang2024ldh,Hu2024gte,Zhang2024tta,Hu2023hdp,Ruhmann2024ibs} and \citet{Zhang2021utm}, amongst others -- have integrated such information to enhance predictive performance and generalisation. Previous work in similar contexts \citep{Yusuf2025ddp} also demonstrated that incorporating information on weather, terrain, socio-demographics, and built environment improved the modelling of stop-level passenger counts.

In the current framework, information from multiple external sources is integrated at separate levels. Weather conditions and temporal information are classified as trip-level context and processed through a dedicated context encoding pathway, while natural terrain and urban characteristics are associated with individual stops and encoded as graph-based embeddings. This design aims to exploit the different roles that these factors play in shaping mobility patterns at different levels of granularity.

\section{Data Pipeline and Feature Engineering}
\label{sec:m-data}

\begin{figure}[tb!]
    \centering \includegraphics[width=\textwidth]{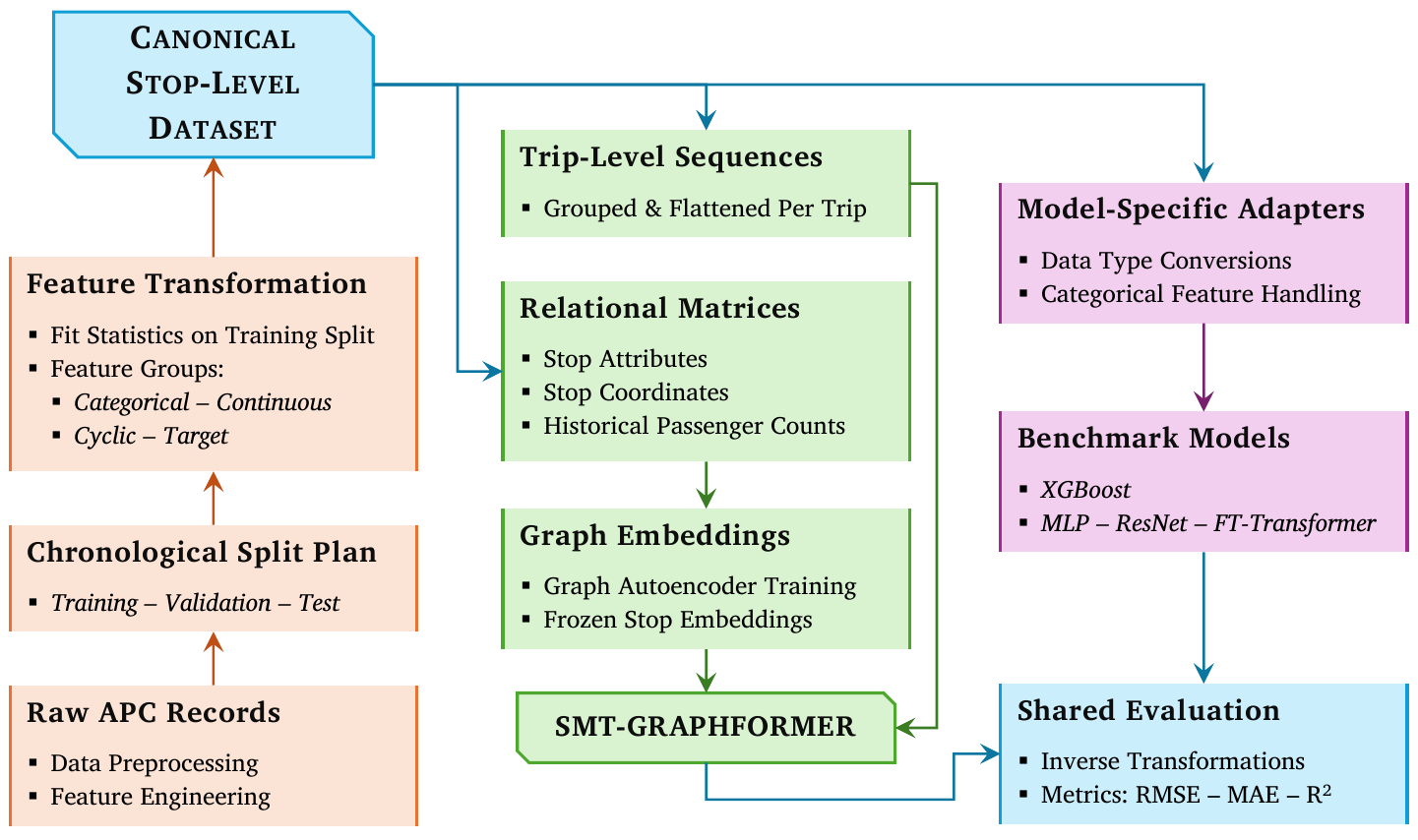}
    \caption{Overview of the data preprocessing and feature engineering pipeline, showing the flow from raw stop-level APC records to the canonical dataset and its transformations for the SMT-GraphFormer and benchmark models.}
    \label{fig:data-pipeline}
\end{figure}

This section describes the pipeline that transforms raw stop-level automated passenger counting (APC) records into the canonical dataset and evaluation protocol shared by \mt{SMT-GraphFormer} and the benchmark models. \autoref{fig:data-pipeline} provides a visual overview of the data pipeline, which can be broadly categorised into four stages: \emph{(i)} initial data preprocessing and feature engineering (orange); \emph{(ii)} canonical stop-level dataset and shared evaluation protocol (blue); \emph{(iii)} trip-level transformation and feature engineering for \mt{SMT-GraphFormer} (green); and \emph{(iv)} minor post-processing for benchmark models (purple).

\subsection{Data Description and Preprocessing}
The raw dataset is a stop-level APC archive provided by AtB, the public transport authority in Trondheim, Norway, covering 31 days in May 2024 across 48 bus lines and 743 bus stops in the Trondheim metropolitan area. Each bus ``line'' is defined as a unique route with a fixed sequence of stops, and each ``trip'' is a single traversal of that route on a given day, with stop visits ordered by their ``stop sequence'' index.
The same stop name may correspond to multiple stop identifiers due to different platforms or travel directions at the same physical location. For example, a stop on a straight road typically maps to two identifiers (one per direction), while a junction stop may correspond to up to four identifiers. This many-to-one mapping results in 1,291 unique stop identifiers across the 743 unique stop names in the dataset.

Each record in the dataset represents a single stop visit within a bus trip and contains 21 fields that cover: identifiers for the line, trip, and stop; scheduled and actual arrival and departure times; stop name and geographic coordinates; binary quality flags for APC sensor activity and extreme value detection; and boarding and alighting passenger counts.
Throughout this work, the term ``stop'' refers to a unique stop identifier, the granularity at which all feature engineering, external data linkage, and relational modelling are performed; stop names are used solely when aggregating transfer connectivity across different lines. The preprocessing stage then applies deterministic rules to handle missing values, enforce type consistency, engineer derivative features, and filter records on quality criteria, as detailed below.

\paragraph{Missing Value Imputation:}
Temporal fields with missing values are filled from their corresponding scheduled arrival or actual departure counterparts. The APC activity flag defaults to \mt{True}, while its extreme value flag defaults to \mt{False}. Missing passenger count values are imputed with zero. Missing bus type information is filled with ``No Information'', except for the tram service on Line~9 where ``Streetcar/Tram'' is assigned.

\paragraph{Derivative Features:}
Since raw trip numbers are reused across different days, a unique trip identifier is constructed by concatenating the date with the trip number to ensure that each trip is uniquely identifiable across the entire dataset. This is crucial for maintaining the integrity of trip-level sequences during data processing and model training. Several features are then computed from the raw fields to capture operational metrics and contextual information.
The dwell time at each stop is calculated as the difference between actual departure and actual arrival times, while the delay is computed as the non-negative deviation from the scheduled arrival time, with negative values clipped to zero to reflect only late arrivals.
Transfer connectivity is quantified by counting the number of other bus lines that serve the same stop name, and a binary label distinguishes transfer stops (those served by multiple lines) from ordinary stops (those served by a single line).
Temporal metadata including day of week, workday and holiday flags, and departure hour are derived from the date and scheduled departure time using the Norwegian public holidays calendar to capture temporal patterns in ridership and operations.

\paragraph{Quality Filtering:}
The intended sequence modelling approach requires complete stop-level trajectories for each trip, so the dataset is filtered to retain only trips with complete records across all stops in their route.
Entire trips (grouped by their unique trip identifier) are removed if: \emph{(i)} the APC sensor was inactive on any stop, \emph{(ii)} extreme passenger count values were flagged, \emph{(iii)} negative boarding or alighting counts were recorded, or \emph{(iv)} the observed set of stop sequences did not match the expected range $\{0, 1, \ldots, n\}$ where $n$ is the last stop sequence index.
The filtered dataset contains 2,218,467 stop-level records spanning 90,329 unique trips across the 31-day study period.

\begin{figure}[tb!]
    \centering
    \begin{minipage}{\textwidth}
        \centering \includegraphics[width=\linewidth]{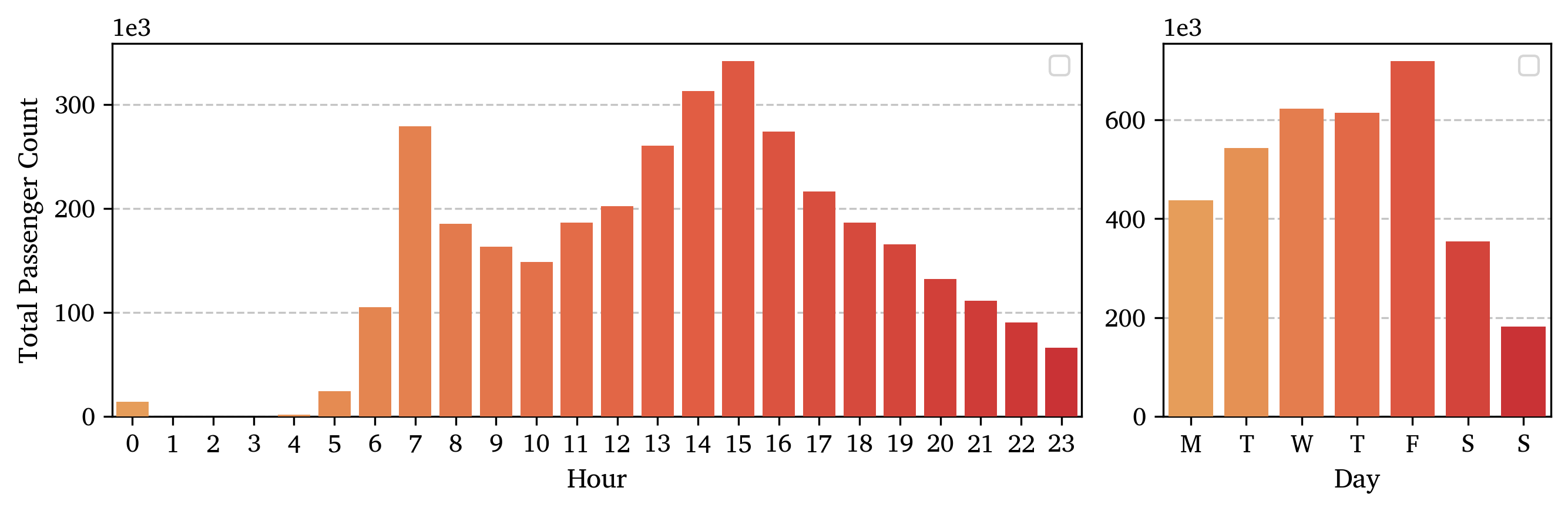}
        \caption{Temporal distribution of passenger counts across the dataset, aggregated by hour of day (left) and day of week (right).}
        \label{fig:overview-hour-day}
    \end{minipage}

    \vspace{\baselineskip}
    \begin{minipage}{\textwidth}
        \centering \includegraphics[width=\linewidth]{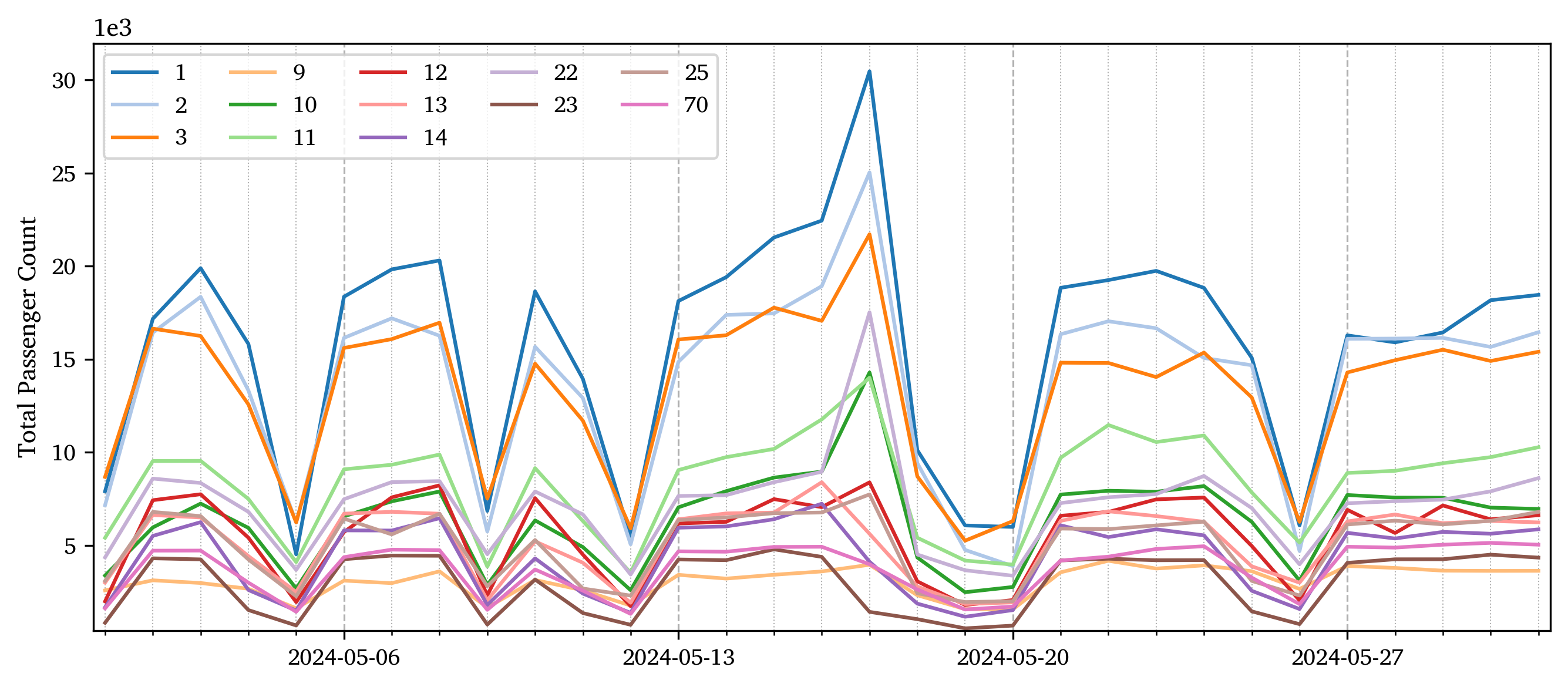}
        \caption{Total daily passenger counts for the busiest lines (85\% of total ridership) across the study period, showing the temporal ridership patterns of individual lines.}
        \label{fig:overview-month-line}
    \end{minipage}
\end{figure}

\paragraph{Exploratory Analysis:}
\autoref{fig:overview-hour-day} shows the temporal distribution of aggregated ridership (boarding counts) across the dataset. The hourly distribution exhibits the expected bimodal pattern with morning and afternoon peaks, while the daily distribution shows increasing ridership across the week with a drop-off on weekends.
\autoref{fig:overview-month-line} further disaggregates the daily ridership across the study period by line, showing the temporal ridership patterns of the busiest lines that together account for 85\% of total ridership. Some lines exhibit pronounced weekday peaks, while others maintain more consistent ridership across the week, underscoring the importance of modelling line-specific dynamics in addition to overall temporal trends.

\subsection{Feature Engineering}
The filtered stop-level dataset encodes the core operational and ridership information for each stop visit, along with relevant temporal metadata. To capture the effect of extrinsic factors on passenger demand and operational performance, the dataset is further enriched with engineered features derived from external data sources, as detailed below.

\paragraph{Natural Terrain:}
The elevation at each stop is obtained from the Copernicus EU-DEM based on its geographic coordinates, with negative values clipped to zero. The elevation data (in meters) is then used to derive several features that characterise the topographic context of each stop and the elevation profile along each trip.
For each consecutive pair of stops $i$ and $i+1$ within a trip, the pipeline computes: \emph{(i)} the straight-line inter-stop distance $d_{i}$ in meters, \emph{(ii)} the signed elevation change $\Delta e_{i} = e_{i+1} - e_{i}$, \emph{(iii)} the absolute elevation change $|\Delta e_{i}|$, \emph{(iv)} a categorical elevation type (\mt{uphill} if $\Delta e_i > 3$\,m, \mt{downhill} if $\Delta e_i < -3$\,m, \mt{flat} otherwise), and \emph{(v)} the slope gradient $\alpha_{i} = \arctan(|\Delta e_{i}| / d_{i})$. For the last stop in each trip, elevation change and inter-stop distance are set to zero.

\paragraph{Urban Characteristics:}
To capture the influence of the socio-demographics and built environment on transit dynamics, each stop is enriched with features describing its urban context, based on official statistics from Statistics Norway (SSB).
Four datasets for the year 2024 were integrated: \emph{(i)} business statistics recording the number of businesses and employees \citep{SSB07091}, \emph{(ii)} resident population \citep{SSB04861}, \emph{(iii)} total dwellings \citep{SSB06265}, and \emph{(iv)} building counts classified according to function \citep{SSB03158}.
These statistics are provided in a grid format at 250\,m resolution, with each cell containing aggregated values for the corresponding features. These grid-based features are linked to each stop by averaging the values of all grid cells within a 250\,m radius of the stop's coordinates, thus capturing the local urban context around each stop while smoothing out small-scale variations.

\paragraph{Weather Conditions:}
Weather conditions can significantly influence individual travel mode choices (public transit \it{vs.} private vehicles \it{vs.} active mobility modes) \citep{Wei2022iti}, and can also cause operational disruptions leading to delays and cancellations \citep{Ngo2024tio}.
To capture these effects, historical daily weather observations for Trondheim covering the study period are sourced from \citet{VisualCrossing2024}. The available weather features were filtered to exclude those with: \emph{(i)} low relevance to public transit, \emph{(ii)} high correlation with other features, or \emph{(iii)} potential for temporal leakage.
The retained features include: daily minimum and maximum temperature, feels-like temperature, maximum wind gust speed, and a categorical weather condition label. These weather features are merged with the stop-level dataset on the date field, so that all stops on a given day share the same weather values.

\subsection{Canonical Stop-Level Dataset}
The preprocessing and feature engineering steps produce a comprehensive canonical stop-level dataset shared by all downstream models. The final feature set (40 in total) is organised into six functional groups: core transit features (11), temporal metadata (4), urban characteristics (10), natural terrain (6), weather conditions (5), and target variables (4). Each feature is described in detail in \autoref{app:feature-details}, including its original form, the transformations applied to it, and how it is integrated into the \mt{SMT-GraphFormer} architecture.

\paragraph{Temporal Split Strategy:}
The temporal split strategy used in this study is designed to ensure generalisation to unseen temporal contexts, reflecting real-world forecasting scenarios where models are trained on historical data and deployed to predict future demand and performance.
The May 2024 dataset is partitioned chronologically by date into training (1--21 May; 21 days; 57,360 trips), validation (22--26 May; 5 days; 14,685 trips), and test (27--31 May; 5 days; 18,284 trips) subsets, yielding an approximate 68:16:16 split by day and a 63:16:20 split by trip count. The unequal trip distribution reflects higher average daily trip volumes in the test period relative to the validation period.
The four Norwegian public holidays in May 2024 fall within the training period, while the validation and test splits reflect normal ridership and operating conditions with no holidays present.

\paragraph{Normalisation and Encoding:}
The canonical features can be categorised into four types: categorical (10), continuous (24), cyclic (2), and target variables (4). The distribution-dependent transformations applied to the continuous features and target variables are fitted exclusively on the training subset to prevent data leakage, while categorical features are encoded using mappings derived from the full dataset to ensure coverage of all categories.
The categorical features are mapped to integer codes using schema-defined mappings where applicable or order-of-appearance indices otherwise. The continuous features are z-score standardised, and the cyclic features are encoded as sine--cosine pairs to avoid discontinuities at temporal boundaries.
Each target variable undergoes a three-step transformation: negative values are clipped to zero, a $\log(1+x)$ transform compresses the right-skewed distributions, and the result is z-score standardised. These transformations define the common target space used by all models during training and prediction.

\paragraph{Evaluation Protocol:}
Evaluation is performed after restoring predictions and transformed ground-truth values to their original units. The inverse target transformation first undoes the z-score standardisation, then reverses the log$(1+x)$ compression, and finally clips values to a minimum of zero, with predictions also capped at the training-set maximum.
This common post-processing is applied to all models before computing three metrics per target variable and data split. These are the mean absolute difference between predictions and ground truth (MAE), the square root of the mean squared difference (RMSE), and the coefficient of determination ($R^2$), which measures explained variance relative to a mean-value baseline. The metrics are computed as follows:

\begin{equation}
    \text{MAE} = \frac{1}{n}\sum_{i}\lvert y_i - \hat{y}_i \rvert  \qquad\big|\qquad
    \text{RMSE} = \sqrt{\frac{1}{n}\sum_{i}(y_i - \hat{y}_i)^2}  \qquad\big|\qquad
    R^2 = 1 - \frac{\sum_{i}(y_i - \hat{y}_i)^2}{\sum_{i}(y_i - \bar{y})^2}
\end{equation}

\noindent where $y_i$ are ground-truth values, $\hat{y}_i$ predictions, and $\bar{y}$ the mean of ground-truth values. The same evaluation pipeline is applied to all models to ensure consistency.

\subsection{Post-Processing for SMT-GraphFormer}
The canonical dataset is used directly by the benchmark models, while \mt{SMT-GraphFormer} applies further operations to adapt the data for sequence modelling and extract relational information for the graph autoencoder.

\paragraph{Trip-Level Transformation:}
Three dummy stop identifiers are reserved for the special tokens used by the sequence model: $\langle$SOS$\rangle$ (start of sequence), $\langle$EOS$\rangle$ (end of sequence), and $\langle$PAD$\rangle$ (padding). Their corresponding feature values are set to zero for all features.
The stop-level dataset is then reshaped into trip-level sequences, where each record is partitioned into trip-level features (those that are constant across all stops in a trip) and stop-level features (those that vary across stops). Each stop-level feature is then flattened into an ordered sequence of per-stop values, bracketed by its corresponding start-of-sequence and end-of-sequence values, and padded to the maximum observed stop sequence length across all trips.
Trip-level features include line identifier, bus type, temporal metadata, and weather information, while stop-level features include stop identifier, scheduled arrival time, stop attributes, and target variables.

\paragraph{Stop Attributes:}
To capture the intrinsic characteristics of each stop and their influence on passenger demand and operational performance, a per-stop feature matrix is constructed from the 10 urban characteristics, 6 natural terrain, and 2 transfer connectivity features. This matrix has dimensions $1{,}294 \times 18$, covering the 1,291 stop identifiers plus the three reserved positions for the special tokens, with all special-token entries set to zero.

\paragraph{Relational Matrices:}
To create a compact representation of the transit network and encode the relational structure of its elements, six dense stop-to-stop similarity matrices are constructed from the static stop attributes and their historical demand profiles.
Each matrix has dimensions $1{,}294 \times 1{,}294$, covering the 1,291 stop identifiers plus the three reserved positions for the special tokens, with all special-token and diagonal entries set to zero. The matrices are constructed from static stop attributes and historical demand profiles as follows:
\emph{(i)} geographic proximity is captured by converting pairwise Euclidean distances to similarities via a Gaussian radial basis function; \emph{(ii)} attribute similarity is computed as the cosine similarity of L2-normalised feature vectors across 18 stop attributes; and \emph{(iii)} mobility pattern similarity is encoded using the Pearson correlation between each pair of stop-level temporal profiles derived from the training subset. Four pattern variants are computed from boarding and alighting counts at both hourly and daily resolutions.
These matrices serve as input to the graph autoencoder, which compresses the multi-relational structure into fixed per-stop embeddings that are consumed by the transformer.

\section{SMT-GraphFormer Architecture and Evaluation}
\label{sec:m-model}
This section describes the \mt{SMT-GraphFormer} architecture, the components that adapt the transformer for the transit prediction task, the training and inference protocols, and the benchmark models used for comparison. The model receives the transformed trip-level sequential data and complementary shared artefacts produced from the canonical stop-level dataset and outputs per-stop predictions for all four target variables: boarding, alighting, delay, and dwell.

\begin{figure}[p]
    \centering
    \begin{minipage}{\textwidth}
        \centering \includegraphics[width=0.84\linewidth]{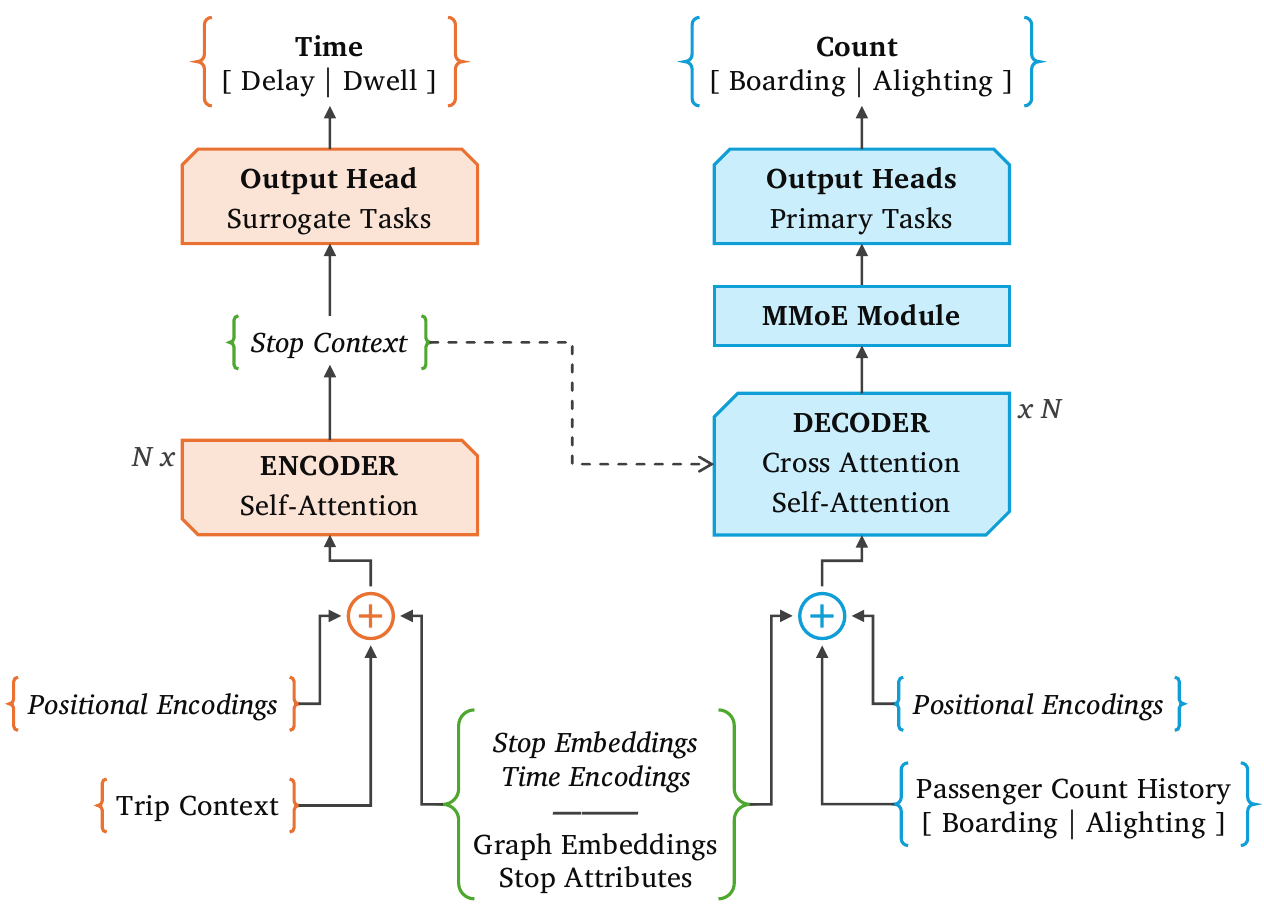}
        \caption{Overview of the SMT-GraphFormer architecture, showing the encoder pathway for surrogate operational predictions and the decoder pathway for autoregressive passenger count predictions. The decoder is conditioned on the encoder output and uses task-specific heads through a multi-gate mixture-of-experts module.}
        \label{fig:smt-architecture}
    \end{minipage}

    \vspace{\baselineskip}
    \begin{minipage}{\textwidth}
        \centering \includegraphics[width=0.54\linewidth]{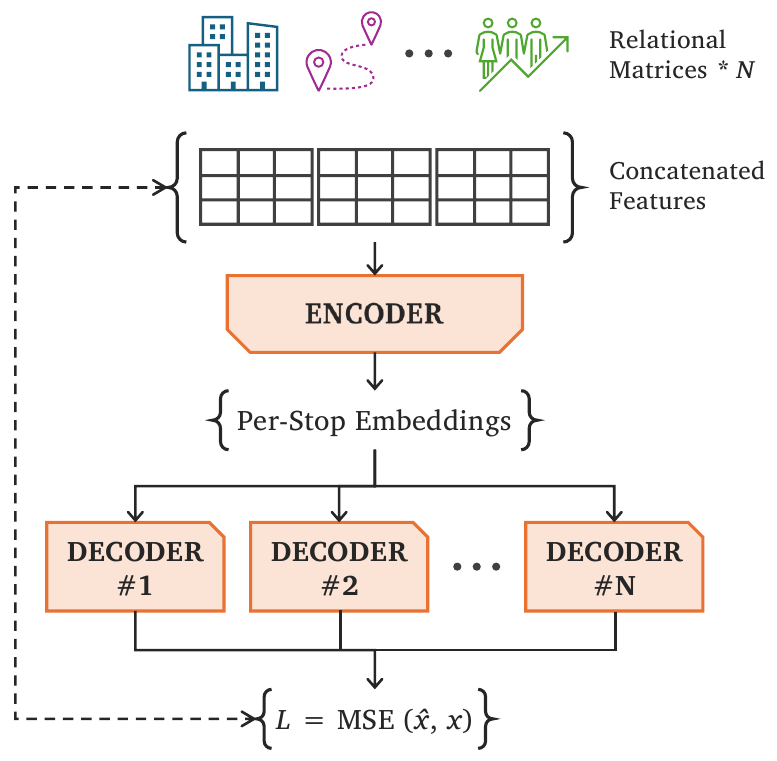}
        \caption{Graph autoencoder architecture, showing the compression of each stop's rows across multiple relational matrices into a dense embedding. Parallel decoder heads reconstruct the corresponding similarity rows independently from the per-stop embedding.}
        \label{fig:graph-autoencoder}
    \end{minipage}
\end{figure}

\subsection{Model Architecture}
The \mt{SMT-GraphFormer} is a modified encoder-decoder transformer adapted from the machine translation paradigm \citep{Vaswani2017aia} to the transit prediction setting, as shown in \autoref{fig:smt-architecture}. Each trip is treated as analogous to a sentence, with bus stops as tokens, and the prediction of passenger counts proceeds sequentially from the first stop to the last.
The encoder ingests the full ordered sequence of stops together with stop-, trip-, and network-level context to produce a condensed representation of the evolving stop context along the trip, while the decoder generates boarding and alighting predictions at each stop conditioned on this context and its own history of previous predictions.

The operational metrics (delay and dwell time) are predicted from the encoder's stop context as surrogate tasks, providing an additional training signal that encourages the model to learn representations informative of the trip's spatiotemporal setting.
In addition, the decoder uses a multi-gate mixture-of-experts (MMoE) module to enable task-specific specialisation for boarding and alighting predictions while sharing computation across the two tasks. The following sections describe each component in detail, starting with the input representation and proceeding through the encoder, decoder, and MMoE module.

The encoder and decoder input representations are constructed from a rich set of features derived from the canonical stop-level dataset, including learned stop embeddings, learned positional encodings, precomputed graph-based embeddings, stop-level attributes, and trip-level context.
A graph autoencoder compresses the multi-relational stop-to-stop similarity matrices into fixed per-stop embeddings, while a context encoder fuses trip-level features into a single context vector.
This comprehensive input representation allows the model to leverage multiple sources of information about the stop sequence, the transit network structure, and the trip context to inform its predictions.

\paragraph{Graph Autoencoder:}
The graph autoencoder (GAE) (\autoref{fig:graph-autoencoder}) compresses the multi-relational stop-to-stop similarity structure into fixed per-stop embeddings as an ``offline'' preprocessing step independent of the main model training. Its input is the stacked similarity tensor of shape $(N_\text{matrices}, N_\text{stops}, N_\text{stops})$, where $N_\text{matrices} = 6$ matrices span $N_\text{stops} = 1{,}294$ stops (1,291 physical stops plus three special tokens).
For each stop, the encoder concatenates its rows across all six matrices into a vector of length $N_\text{stops} \times N_\text{matrices} = 7{,}764$ and passes it through a three-layer feedforward network that compresses the high-dimensional similarity information into dense embeddings of dimension $d_\text{embed}$.
The decoder comprises six parallel heads, one per matrix, each a three-layer network that reconstructs the corresponding similarity row of dimension $N_\text{stops}$.
The GAE is trained by minimising the mean MSE reconstruction loss across the six matrices. After training, the encoder produces the per-stop embedding matrix of shape $(N_\text{stops}, d_\text{embed})$, which is detached and passed to the main model as a frozen buffer.

\paragraph{Context Encoder:}
The context encoder fuses trip-level scalar features into a fixed-size context vector that is broadcast across all stop positions in the encoder pathway. It receives 15 features per trip covering operational, temporal, and weather information. Six are categorical, while the remaining nine comprise five continuous features and four cyclic encodings.
Each categorical feature is mapped through a learned embedding of dimension $d_\text{context}$ and the resulting embeddings are directly concatenated with the z-scored continuous features and cyclic encodings to form a single context vector of length 105. This vector is then passed through LayerNorm and a two-layer MLP that projects it to \mt{SMT-GraphFormer}'s embedding dimension $d_\text{embed}$.

\paragraph{Input Representation:}
The model constructs its input representation by summing six components at each stop position, with the encoder and decoder differing only in their additional pathway-specific signals.
Two-layer feedforward adapter networks are applied across the model to project input features and output predictions between their native dimensions and the model's embedding dimension $d_\text{embed}$ where necessary.
Learnable gates allow the model to modulate the influence of each component during training, with the exception of the context encoder contribution which is controlled by a fixed gate that can be disabled for ablation.

A learned embedding layer maps each of the 1,294 stop token indices (1,291 physical stops plus three special tokens) to a dense vector of dimension $d_\text{embed}$. The z-scored scheduled arrival time at each stop is projected from a scalar to $d_\text{embed}$ through an adapter.
The precomputed GAE embeddings are registered as a frozen buffer and looked up at each position, then projected through an adapter and scaled by a learnable scalar gate. Although the GAE embeddings are already at the model's embedding dimension, the adapter allows the model to learn its own transformation of the independently trained GAE space.
The stop attributes matrix of shape $(N_\text{stops}, 18)$ is looked up at each position, projected to $d_\text{embed}$, and scaled by a learnable gate.

These four components are shared between the encoder and decoder pathways, along with two additional components that differ between the pathways. Separate learned positional embedding tables for the encoder and decoder map the stop positions within the trip to the embedding dimension.
For the encoder, the context encoder output is broadcast-added to all stop positions, with a fixed gate (configurable for ablation) controlling this contribution. For the decoder, the ground-truth (during training) or previously predicted (during inference) boarding and alighting values at each position are projected to $d_\text{embed}$ through an adapter.
The final input representation at each position is the sum of all components, yielding a tensor of shape $(B, T, d_\text{embed})$ where $B$ is the batch size and $T$ the contextualised trip length (including special and padding tokens).

\paragraph{Encoder and Decoder:}
The sequence backbone follows a standard pre-norm encoder-decoder transformer structure, with multi-head attention, residual connections, position-wise feedforward networks, and final layer normalisation in both pathways.
The encoder processes the full stop sequence, using a key-padding mask to suppress padding positions, and produces contextual stop representations of shape $(B, T, d_\text{embed})$. These representations serve as memory for decoder cross-attention and as inputs to an encoder-side prediction head for the surrogate operational targets (delay and dwell time).
The decoder receives a shifted stop sequence and produces autoregressive decoder states of shape $(B, T, d_\text{embed})$, conditioned on the encoder memory through cross-attention. These states do not directly represent passenger-count predictions, but are passed to the MMoE module for task-specific processing before the boarding and alighting heads. The decoder's causal self-attention ensures that each position can only attend to information from the same or previous stops, while cross-attention allows the decoder states to leverage the full encoder context.

\begin{figure}[tb!]
    \centering \includegraphics[width=0.65\textwidth]{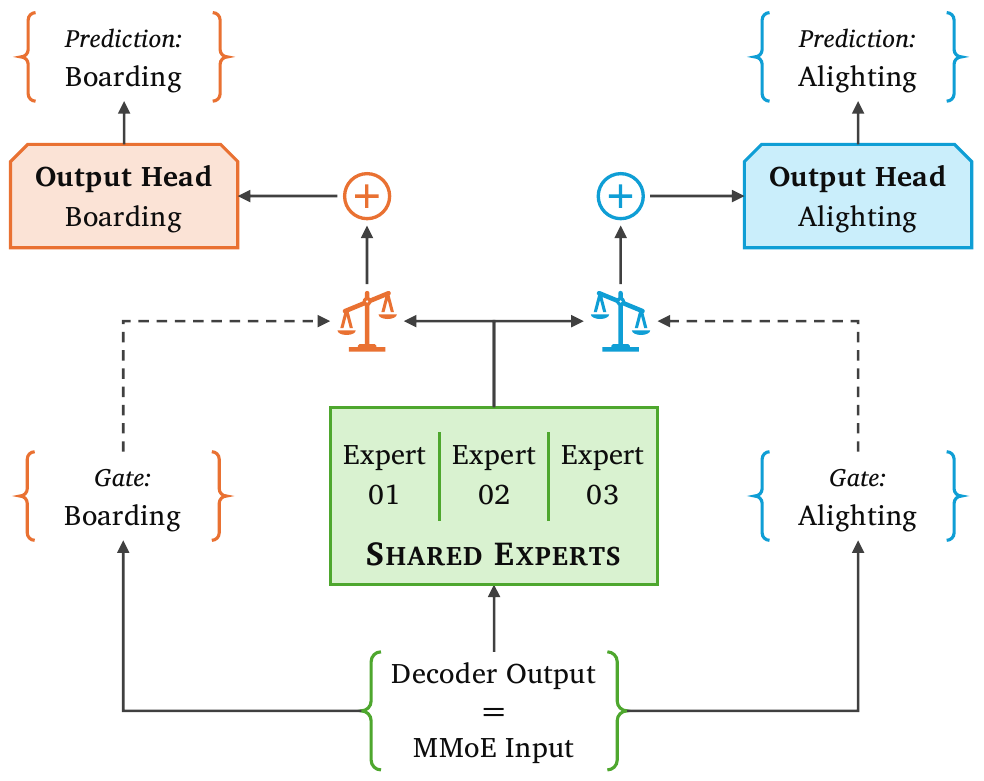}
    \caption{Multi-gate mixture-of-experts module. Each expert processes the decoder output independently, while task-specific gates produce weights over experts to yield one specialised representation per task. \it{Note: Adapted from \citet{Hu2024gte}.}}
    \label{fig:mmoe-module}
\end{figure}

\paragraph{MMoE:}
The MMoE module (\autoref{fig:mmoe-module}) sits between the decoder output and the passenger-count prediction heads, enabling task-specific specialisation while sharing computation across the two primary targets (boarding and alighting). It comprises $N_\text{expert}$ expert networks and two task-specific gates.
Each expert is a two-layer feedforward network of fixed width $d_\text{embed}$ that processes the decoder output independently to learn different transformations of the decoder states that may be more or less relevant for each task.
Each task gate is a learned linear projection to $N_\text{expert}$ followed by softmax, producing a weight vector over experts. The task output is the corresponding weighted sum applied to the expert outputs, yielding one specialised representation per task of shape $(B, T, d_\text{embed})$.
Two independent decoder-side prediction heads then map these representations to scalar boarding and alighting predictions at each position.

\subsection{Model Training and Inference}

\paragraph{Loss Functions:}
The total training loss $\mathcal{L}$ combines a primary loss for passenger-count predictions from the decoder with a surrogate loss for operational metric predictions from the encoder. Both terms are computed as mean squared error (MSE) over non-padding positions. The surrogate task provides an auxiliary training signal that encourages the encoder to learn representations informative of operational performance, which may indirectly benefit the primary passenger-count predictions.

\begin{equation}
    \begin{aligned}
        \mathcal{L}_\text{primary}   & = \text{MSE}\,\bigl(\hat{b}[\mathbf{m}],\; b[\mathbf{m}]\bigr) + \text{MSE}\,\bigl(\hat{a}[\mathbf{m}],\; a[\mathbf{m}]\bigr) \\
        \mathcal{L}_\text{surrogate} & = \text{MSE}\,\bigl(\hat{s}[\mathbf{m}],\; s[\mathbf{m}]\bigr)                                                                \\
        \mathcal{L}                  & = \mathcal{L}_\text{primary} + w_\text{surrogate} \cdot \mathcal{L}_\text{surrogate}
    \end{aligned}
\end{equation}

\noindent where $\hat{b}$ and $\hat{a}$ are predicted boarding and alighting counts, $b$ and $a$ the ground-truth values, $\mathbf{m}$ the non-padding mask, and $s = [\text{delay}, \text{dwell}]^{\top}$ the surrogate targets. The weight $w_\text{surrogate}$ controls the relative importance of the surrogate task and can be tuned or set to zero for ablation.

\paragraph{Scheduled Sampling:}
Under standard teacher-forced training, the decoder always receives the ground-truth passenger count history as input. During inference, the model must feed its own predictions back for subsequent positions, creating a distribution mismatch known as exposure bias that can lead to error accumulation along the trip.
To mitigate this, the model is also trained with a scheduled sampling variant in which the decoder progressively replaces ground-truth inputs with its own predictions during training, exposing it to its own errors and teaching it to recover from them.
The replacement probability $p$ remains at zero throughout the warmup phase and then increases linearly to a maximum $p_\text{max}$ over the remaining iterations. At each step where $p > 0$, a teacher-forced forward pass first obtains predictions at all positions without gradient tracking. A Bernoulli mask with probability $p$ then determines which positions receive the model's detached predictions in place of ground truth, and a second forward pass with gradient tracking computes the loss from these mixed inputs.

\paragraph{Inference Protocols:}
Two inference protocols are supported. In the \it{teacher-forced} protocol, the full ground-truth boarding and alighting sequences are provided as decoder inputs in a single forward pass. In the \it{autoregressive} protocol, the decoder generates boarding and alighting predictions one stop at a time, initialised with fill values for the first position ($\langle$SOS$\rangle$). At each step, the predicted passenger counts are fed back as decoder inputs for subsequent positions. The encoder's stop context is computed once and reused across all decoding steps.

\subsection{Benchmark Models}
The benchmarks are chosen to represent the strongest non-sequential alternatives. They receive the same input features and predict the same four targets, but treat each stop visit as an independent tabular record rather than as an explicit position within a trip sequence. This choice isolates the contribution of the \mt{SMT-GraphFormer}'s sequential formulation and graph-based components from the value of the enriched feature set alone. The benchmarks share the same canonical transformed stop-level dataset and evaluation protocol as the main model. The complete training and hyperparameter configurations for each model are provided in \autoref{app:model-configs}.

\subsubsection{XGBoost}
XGBoost (Extreme Gradient Boosting) \citep{Chen2016xas} is a popular gradient-boosted decision tree implementation which remains a strong baseline for tabular modelling tasks. It builds trees sequentially, each learning to correct the residual errors of the preceding ensemble using gradient-based optimisation.
XGBoost represents a strong reference point for this data and has been shown to capture non-linear interactions between stop-level attributes despite treating each record independently without an explicit relational inductive bias \citep{Yusuf2025ddp}.
For this task, a meta-regressor trains a dedicated set of trees per target within a single coordinated boosting process, with iteration count and early stopping applied jointly across all targets. The native handling for categorical features is enabled.

\subsubsection{RTDL Models}
Three deep learning (DL) architectures from the RTDL (Research on Tabular Deep Learning) benchmark suite \citep{Gorishniy2021rdl} are selected that together span the main families of tabular DL models. These models serve as a well-calibrated complement to XGBoost for evaluating the \mt{SMT-GraphFormer} against both simple and sophisticated tabular alternatives.

\paragraph{MLP:}
The multi-layer perceptron (MLP) is the simplest deep baseline, consisting of stacked fully connected blocks with ReLU activations and dropout. It receives a flat input vector in which categorical features are one-hot encoded, and learns feature interactions purely through depth and width.

\paragraph{ResNet:}
Here, the residual network (ResNet) paradigm is adapted to tabular data by applying an input projection followed by a stack of residual blocks with skip connections. The skip connections provide shorter gradient paths from input to output, enabling effective training of deeper networks than a plain MLP. \citet{Gorishniy2021rdl} found ResNet to be a strong tabular baseline that is often missing from tabular DL comparisons. Like MLP, it operates on a flat input vector with one-hot-encoded categorical features.

\paragraph{FT-Transformer:}
The Feature Tokeniser Transformer (FT-Transformer) was introduced by \citet{Gorishniy2021rdl} and has shown competitive performance across several tabular modelling tasks. FT-Transformer treats all input features (categorical and continuous) as tokens, transforms them into embeddings, and applies multi-head self-attention to learn pairwise interactions across the full feature set. Categorical features are embedded through learned lookup tables, while continuous features are projected through feedforward adapters. This approach avoids the high-dimensional sparse one-hot representations required by the MLP and ResNet for high-cardinality features such as the stop identifier. FT-Transformer is the most architecturally sophisticated of the tabular DL models and shares the same transformer heritage as the \mt{SMT-GraphFormer}, while remaining a purely stop-level model with no sequential structure. Its inclusion helps to directly test whether a transformer architecture alone is sufficient to capture the necessary feature interactions for this task.

\section{Results and Discussion}
\label{sec:results}
This section evaluates the \mt{SMT-GraphFormer} against the stop-level benchmarks, followed by ablation studies examining individual architectural components and an analysis of the autoregressive inference gap. The discussion focuses on the test-set $R^2$ as the primary performance metric, since its scale-free nature allows for direct comparison across targets with different units and magnitudes (passenger counts versus seconds). Unless otherwise noted, \mt{SMT-GraphFormer} results refer to teacher-forced evaluation, which is directly comparable to the stop-level benchmarks that do not have an autoregressive component.

\subsection{Experimental Setup}
\mt{SMT-GraphFormer} is optimised with AdamW using a learning rate of $3 \times 10^{-4}$ and weight decay of $10^{-2}$, with gradients clipped to a maximum norm of 1.0. The learning rate follows a linear warmup from 1\% of the base rate over the first 7.5\% of training iterations, then cosine annealing to $10^{-6}$ over the remainder. Training proceeds for a maximum of 40 epochs with a batch size of 256, with 10\% dropout applied throughout the model. Detailed architecture and training configurations for all models are provided in \autoref{app:model-configs}.

All experiments were conducted on a Linux HPC cluster node equipped with an NVIDIA V100 GPU (16\,GB VRAM), 16-core CPU, and 64\,GB of system RAM. The software environment comprised Python~3.12 and PyTorch~2.9.1+cu128. XGBoost was trained using the \mt{xgboost} library, and the RTDL models were trained using code adapted from the official \mt{rtdl_revisiting_models} package. To ensure reproducibility, random seeds were fixed for Python, NumPy, and PyTorch, and PyTorch's deterministic execution mode was enabled across all runs. The full codebase is publicly available at \url{https://github.com/Outsiders17711/SMTGraphFormer}.

\begin{figure}[tb!]
    \centering \includegraphics[width=\textwidth]{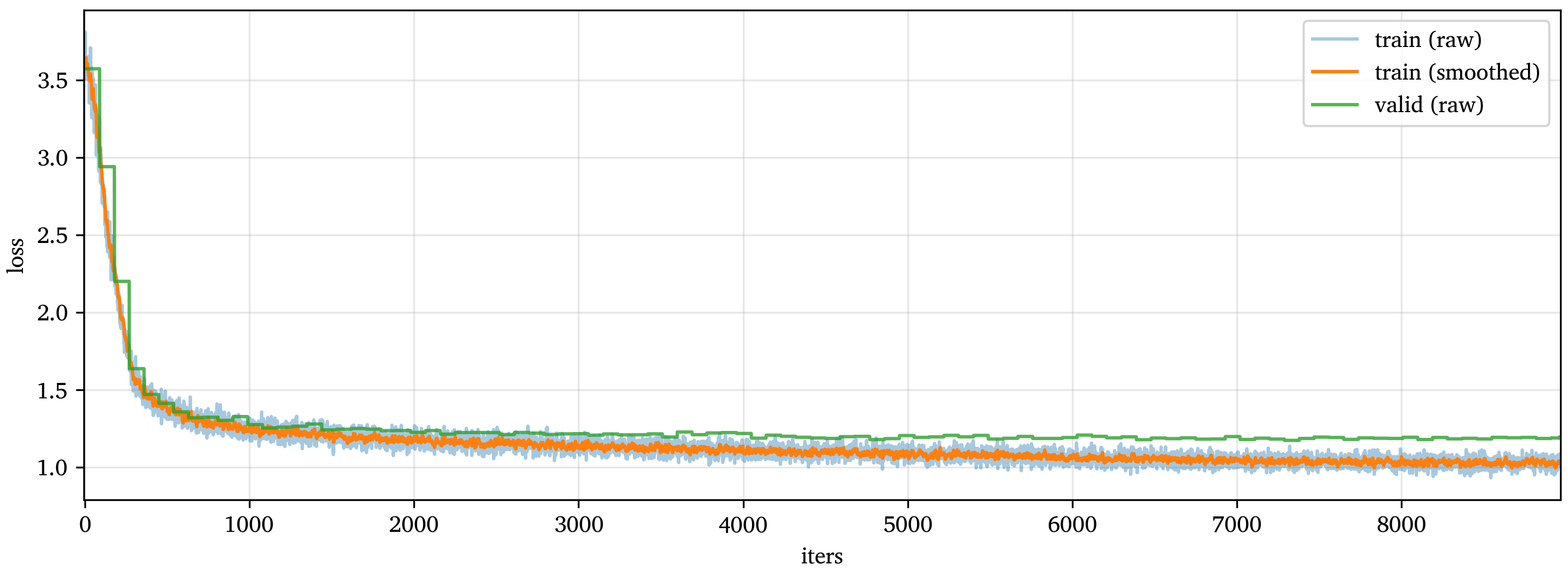}
    \caption{Training curves for the SMT-GraphFormer baseline, showing training and validation loss across iterations for the combined primary and surrogate objectives. \it{Note: The validation loss is computed at sparse intervals, which results in the stepped appearance of the validation curve.}}
    \label{fig:training-history}
\end{figure}

\subsection{Baseline Performance}
\autoref{tab:smt-baseline} presents the detailed per-target performance of \mt{SMT-GraphFormer} across all three data splits under teacher-forced evaluation. The complementary training curves in \autoref{fig:training-history} show smooth convergence of the combined loss with minimal divergence between training and validation losses across iterations. While the training loss continues to decrease, the validation loss stopped improving after approximately 33 epochs, suggesting that the model has reached its optimal generalisation performance within the current architecture and training configuration.

The most notable performance is on the primary targets: boarding achieves a test $R^2$ of 0.550 and alighting reaches 0.715, the latter benefiting substantially from the sequential formulation in which ground-truth boarding history at preceding stops provides a strong signal for downstream alighting. The operational targets are harder to predict, likely reflecting the dominance of real-world stochasticity that the input features cannot capture. This appears most clearly in delay, which achieves a test $R^2$ of only 0.186, while dwell performs better at 0.346, consistent with its closer coupling to passenger volumes rather than external traffic conditions.

The train--test $R^2$ gaps are moderate for boarding (0.101) and alighting (0.068), indicating reasonable generalisation to unseen data without overfitting. There is a larger gap for delay of 0.214, which was even more pronounced on the validation set (0.271), suggesting that the distinct temporal character of the splits described in \autoref{sec:m-data} may have a significant impact on delay prediction. On the other hand, dwell shows more consistent performance across splits with a gap of only 0.009, suggesting that its explained variance is stable and generalisable.

\begin{table}[pos=tb!]
    \centering \renewcommand{\arraystretch}{1.15}
    \caption{SMT-GraphFormer baseline performance across training, validation, and test splits under teacher-forced evaluation. \it{Note: Boarding and alighting are unitless passenger counts, while delay and dwell are in seconds.}}
    \label{tab:smt-baseline}
    \begin{tabular*}{0.60\textwidth}{l@{\extracolsep{\fill}}l@{\qquad}|ccc}
        \toprule
        \bt{Target} & \bt{Split} & \bt{RMSE} & \bt{MAE} & \bm{R^2} \\
        \midrule
        Boarding & Training & 1.901 & 0.887 & 0.651 \\
        & Validation & 1.978 & 0.940 & 0.554 \\
        & Test & 1.861 & 0.892 & 0.550 \\
        \midrule
        Alighting & Training & 1.538 & 0.732 & 0.783 \\
        & Validation & 1.597 & 0.786 & 0.716 \\
        & Test & 1.505 & 0.743 & 0.715 \\
        \midrule
        Delay & Training & 121.590 & 66.583 & 0.400 \\
        & Validation & 130.928 & 76.026 & 0.129 \\
        & Test & 116.498 & 70.958 & 0.186 \\
        \midrule
        Dwell & Training & 34.213 & 8.914 & 0.355 \\
        & Validation & 34.430 & 9.236 & 0.337 \\
        & Test & 33.739 & 9.109 & 0.346 \\
        \bottomrule
    \end{tabular*}
\end{table}

\subsection{Benchmark Comparison}
\autoref{tab:benchmark-comparison} compares the test-set $R^2$ scores of the \mt{SMT-GraphFormer} baseline against the stop-level benchmarks, with detailed per-split results provided in \autoref{app:benchmark-details}. The benchmark models are ordered by increasing architectural complexity, but the results show that complexity does not translate into a monotonic improvement in performance, as also noted by \citet{Gorishniy2021rdl} and \citet{Yusuf2025ddp}.
XGBoost remains competitive, slightly outperforming MLP and ResNet on all targets. ResNet does not improve on MLP, indicating that increased depth alone does not guarantee better performance. FT-Transformer remains the strongest stop-level benchmark overall, suggesting that its per-feature tokenisation and self-attention mechanism provide a better inductive bias for tabular transit data.

\begin{table}[pos=tb!]
    \centering \renewcommand{\arraystretch}{1.15}
    \caption{Test-set $R^2$ comparison across all models. \it{Note: The best-performing model for each target is highlighted in bold.}}
    \label{tab:benchmark-comparison}
    \begin{tabular*}{0.65\textwidth}{l@{\extracolsep{\fill}}@{\qquad}|cccc}
        \toprule
        \bt{Model} & \bt{Boarding} & \bt{Alighting} & \bt{Delay} & \bt{Dwell} \\
        \midrule
        XGBoost & 0.487 & 0.449 & 0.115 & 0.311 \\
        MLP & 0.484 & 0.440 & 0.096 & 0.309 \\
        ResNet & 0.486 & 0.439 & 0.046 & 0.301 \\
        FT-Transformer & 0.511 & 0.476 & 0.147 & 0.300 \\
        SMT-GraphFormer & \bt{0.550} & \bt{0.715} & \bt{0.186} & \bt{0.346} \\
        \bottomrule
    \end{tabular*}
\end{table}

\mt{SMT-GraphFormer} outperforms all stop-level benchmarks on all four targets, improving on FT-Transformer by $+$0.038 on boarding, $+$0.239 on alighting, $+$0.039 on delay, and $+$0.046 on dwell. The dominant gain is on alighting, which is strongly shaped by upstream boarding history that the stop-level FT-Transformer cannot adequately exploit despite also being transformer-based. The more modest gains on boarding, delay, and dwell reflect their weaker dependence on within-trip dynamics: boarding is driven mainly by local stop attributes and temporal context, delay by external traffic conditions, and dwell partly by passenger volumes.

The consistent improvement across all targets suggests that \mt{SMT-GraphFormer}'s sequential formulation, inter-target interactions, and shared representations provide a useful inductive bias beyond the alighting-specific benefit. Coupled with the non-monotonic relationship between architectural complexity and performance observed in the benchmarks, these results underscore the need for adapting DL architectures to the specific structure of transit data rather than assuming that more complex off-the-shelf models will automatically yield better performance.

\subsection{Component Ablations}
The architecture supports systematic ablation of five components. The graph embeddings, stop attributes, and context encoding are disabled by zeroing their respective gates, while surrogate tasks are removed by setting $w_\text{surr} = 0$ in the combined loss. In the surrogate-task ablation, the delay and dwell heads no longer receive a training signal, so those outputs are not evaluated. The MMoE module is replaced with a pass-through that copies the decoder output identically to the decoder-side prediction heads. To isolate the contribution of each component, six ablation configurations are evaluated with all other hyperparameters kept identical to the baseline. \autoref{tab:ablation-results} presents the test-set $R^2$ for each configuration.

\begin{table*}[pos=tb!]
    \centering \renewcommand{\arraystretch}{1.15}
    \caption{Test-set $R^2$ results for SMT-GraphFormer component ablations. Each row disables one component relative to the baseline version; the stripped version disables all optional components simultaneously. \it{Note: Differences from the baseline are shown in parentheses; improvements are marked with $+$ and degradations with $-$; dashes indicate omitted metrics.}}
    \label{tab:ablation-results}
    \begin{tabular*}{0.97\textwidth}{l@{\extracolsep{\fill}}@{\quad}|cccc}
        \toprule
        \bt{Configuration} & \bt{Boarding} & \bt{Alighting} & \bt{Delay} & \bt{Dwell} \\
        \midrule
        Baseline (Full Model) & 0.550 & 0.715 & 0.186 & 0.346 \\
        Stripped (No Optional Components) & 0.555 ($+$0.006) & 0.719 ($+$0.005) & -- & -- \\
        \midrule
        No Graph Embeddings & 0.538 ($-$0.012) & 0.708 ($-$0.006) & 0.199 ($+$0.013) & 0.335 ($-$0.010) \\
        No Stop Attributes & 0.534 ($-$0.016) & 0.705 ($-$0.010) & 0.144 ($-$0.042) & 0.348 ($+$0.002) \\
        No Context Encoding & 0.556 ($+$0.007) & 0.722 ($+$0.007) & 0.200 ($+$0.014) & 0.332 ($-$0.014) \\
        No Surrogate Tasks & 0.542 ($-$0.008) & 0.696 ($-$0.019) & -- & -- \\
        No MMoE & 0.551 ($+$0.001) & 0.712 ($-$0.003) & 0.146 ($-$0.040) & 0.346 ($+$0.001) \\
        \bottomrule
    \end{tabular*}
\end{table*}

The ablations show that stop attributes and graph embeddings are the most consistently useful input components. Removing graph embeddings reduces boarding by 0.012, alighting by 0.006, and dwell by 0.010, while delay improves slightly by 0.013. This suggests that stop topology is more useful for demand-related targets than for delay. Removing stop attributes produces the broadest degradation, especially for delay, which falls by 0.042, indicating that natural terrain and urban characteristics provide local context that is not recoverable from the other input channels. Surrogate supervision also contributes to the primary decoder tasks, since removing it reduces boarding by 0.008 and alighting by 0.019. This suggests that delay and dwell supervision regularises the shared encoder and improves the representations passed to the decoder through cross-attention.

The remaining components have weaker or more mixed effects. Removing context encoding improves boarding, alighting, and delay, but reduces dwell by 0.014, suggesting that the context pathway may duplicate information already available through the graph embeddings and scheduled arrival times. Replacing the MMoE with a pass-through leaves the passenger count targets almost unchanged, while delay falls by 0.040. Since delay is predicted from the encoder-side surrogate head rather than through the MMoE itself, this effect is best interpreted as an indirect interaction within the shared multi-task training objective rather than a direct benefit of expert routing.

The stripped model marginally exceeds the full baseline on boarding and alighting, suggesting some interaction cost among components that are individually useful. However, the full model is the better-balanced configuration for the broader multi-task objective, providing operationally useful delay and dwell predictions alongside competitive passenger-count predictions. This pattern suggests that the issue is less the presence of graph, stop-attribute, or context signals in isolation than how they are injected into the shared representation. More targeted conditioning, such as layer-specific gates for graph embeddings and stop attributes, or moving temporal and line-level context closer to the output heads, may reduce this entanglement while preserving the broader multi-task benefit.

\subsection{Autoregressive Evaluation and Scheduled Sampling}
Autoregressive evaluation tests the decoder under deployment-like conditions, where predicted boarding and alighting values are fed back as inputs at subsequent stop positions. This differs from teacher-forced evaluation, where ground-truth passenger counts are provided as decoder inputs and error accumulation is removed. Delay and dwell are produced by the encoder-side surrogate heads and are therefore not affected by this inference loop. \autoref{tab:autoreg-results} compares the test-set $R^2$ under both evaluation protocols together with a scheduled sampling variant, which changes the training signal by exposing the model to its own predictions during training.

\begin{table}[pos=tb!]
    \centering \renewcommand{\arraystretch}{1.15}
    \caption{Test-set $R^2$ under teacher-forced (TF) and autoregressive (AR) evaluation, including scheduled sampling (SS) as a training variant. \it{Note: Differences in parentheses are relative to the corresponding baseline protocol.}}
    \label{tab:autoreg-results}
    \begin{tabular*}{0.90\textwidth}{l@{\extracolsep{\fill}}@{\quad}|cccc}
        \toprule
        \bt{Evaluation} & \bt{Boarding} & \bt{Alighting} & \bt{Delay} & \bt{Dwell} \\
        \midrule
        Baseline (TF) & \bt{0.550} & \bt{0.715} & \bt{0.186} & \bt{0.346} \\
        Baseline (AR) & 0.515 ($-$0.035) & 0.474 ($-$0.241) & 0.186 & 0.346 \\
        \midrule
        Scheduled Sampling (TF) & 0.550 ($-$0.000) & 0.683 ($-$0.032) & 0.146 ($-$0.040) & 0.339 ($-$0.007) \\
        Scheduled Sampling (AR) & 0.513 ($-$0.002) & 0.473 ($-$0.001) & 0.146 ($-$0.040) & 0.339 ($-$0.007) \\
        \bottomrule
    \end{tabular*}
\end{table}

Switching from teacher-forced to autoregressive evaluation reveals an asymmetric degradation. Boarding drops by only 0.035, while alighting falls by 0.241 to 0.474, almost identical to the FT-Transformer score of 0.476 in \autoref{tab:benchmark-comparison}. The teacher-forced gain from the sequential formulation is therefore largely lost when the model must rely on its own predicted passenger counts. This reflects the different autoregressive dependence of the two passenger-count targets: boarding is driven mainly by local stop characteristics and trip-level context available through cross-attention, whereas alighting depends more directly on upstream boarding history and is more vulnerable to accumulated prediction error.

To investigate whether the gap can be narrowed through exposure-bias mitigation, a scheduled sampling variant was trained in which the replacement probability $p$ ramps linearly from 0 after learning rate warmup to a maximum of 0.5. This replaces some ground-truth decoder inputs with predictions from a gradient-free first pass. The strategy does not improve autoregressive performance, with boarding and alighting changing by only $-$0.002 and $-$0.001 relative to the baseline autoregressive scores. Instead, teacher-forced performance degrades, most notably for alighting ($-$0.032) and delay ($-$0.040). The smaller teacher-forced to autoregressive gap is therefore caused by a weaker teacher-forced model rather than improved autoregressive robustness.

The results suggest that the decoder benefits from a clean passenger-count history under teacher forcing, but is sensitive to errors when that history is replaced by self-generated predictions. Once the autoregressive trajectory becomes noisy, the remaining stop-level and contextual channels are not sufficient to recover the alighting performance observed under teacher forcing. Scheduled sampling introduces this noise during training, but in this configuration it weakens the teacher-forced model without improving autoregressive robustness. Narrowing the gap would therefore require either making the autoregressive pathway more robust to accumulated error or removing the loop by routing encoder-side passenger-count estimates to the decoder.

\section{Conclusion}
\label{sec:conclusion}

This paper presented \mt{SMT-GraphFormer}, a spatiotemporal multi-task graph transformer that frames trip-level public transit prediction as a sequence modelling task. The model uses a modified encoder-decoder transformer enriched with graph-based stop embeddings and trip-level context. An encoder-side prediction head estimates delay and dwell times as surrogate tasks, while a multi-gate mixture-of-experts module produces task-specific decoder representations that feed into the boarding and alighting heads. This architecture is supported by a data pipeline that transforms stop-level transit records into enriched trip-level sequences with terrain, urban, weather, and relational network features. Together, these elements form a horizon-agnostic framework that can produce complete per-stop trajectories for a given line, stop sequence, and trip context, making it suitable for scenario analysis in digital twin environments.

Experimental evaluation on data from the public bus transit system in Trondheim, Norway, confirmed that the sequential formulation yields substantial gains over stop-level benchmark models under teacher-forced evaluation. The model achieved test $R^2$ scores of 0.55 for boarding, 0.71 for alighting, 0.19 for delay, and 0.35 for dwell. The largest gain was observed for alighting, with a $+$0.24 $R^2$ improvement over the best-performing benchmark, alongside smaller but consistent gains on boarding ($+$0.04), delay ($+$0.04), and dwell ($+$0.05). Since alighting is strongly shaped by upstream boarding history, this result supports the central hypothesis that explicit trip-level sequential bias and inter-target dependencies provide a useful inductive bias for transit prediction.

The principal limitations concern the autoregressive inference gap and the handling of trip-level context. In offline deployment, error accumulation under autoregressive decoding substantially degrades alighting performance, and scheduled sampling was not sufficient to close this gap in the tested configuration. This limitation is less severe in online settings where observed passenger counts from previous stops can be used as decoder inputs, but it remains important for fully simulated scenarios. The context encoder also showed mixed effects in the ablation study, suggesting that the current injection pathway may duplicate or interfere with information already available through other channels. Future work should therefore focus on improving robustness under autoregressive inference, redesigning the context pathway, and simplifying the architecture for greater interpretability and efficiency without sacrificing predictive performance.

\section*{Acknowledgement}
This research received funding from the PERSEUS project, part of the European Union's Horizon 2020 research and innovation programme under the Marie Skłodowska-Curie grant agreement No. 101034240. The authors also acknowledge the financial support of MobilitetsLab Stor-Trondheim (MoST), a collaborative initiative for research and development of future-oriented sustainable urban mobility solutions in Norway. The authors are grateful to AtB AS, the public transport authority in Trondheim, for sharing mobility data and insights that informed our analysis.



\clearpage
\appendix

\setcounter{table}{0}
\setcounter{figure}{0}
\renewcommand{\thetable}{A\arabic{table}}
\renewcommand{\thefigure}{A\arabic{figure}}

\renewcommand{\theHtable}{A\arabic{table}}
\renewcommand{\theHfigure}{A\arabic{figure}}

\section{Detailed Feature Descriptions}
\label{app:feature-details}

\newcommand{\dtu}[2]{-- #1 \newline -- #2}
\newcommand{\fcat}[1]{\rule{0pt}{\dimexpr\ht\strutbox+3pt\relax}\bt{#1}}

\begin{table}[pos=H]
    \centering \renewcommand{\arraystretch}{1.25} \footnotesize
    \caption{Canonical stop-level feature descriptions, transformations, and model usage, categorised by core transit features, temporal metadata, urban characteristics, natural terrain, weather conditions, and target variables.}
    \label{tab:feature-details}
    \begin{tabular*}{\textwidth}{p{0.17\textwidth}p{0.78\textwidth}}
        \toprule
        \bt{Feature} & \bt{Description, Transformation, and Usage} \\
        \midrule \midrule
        & \fcat{Core Transit Features (11)} \\[3pt]
        \mt{StopIdentifier} & \dtu{Categorical. The unique identifier for the stop in NeTEx format, serving as the primary key for all stop-level lookups across the dataset.}{Mapped to an integer code and used as an index into a learnable stop embedding table. The resulting stop embedding is shared between the encoder and decoder.} \\
        \midrule
        \mt{StopSequence} & \dtu{Categorical. The ordinal position of the stop within the trip, 0-indexed from the first stop.}{Used directly as a sequence index into a learnable positional encoding table. Separate positional encoding tables are maintained for the encoder and the decoder.} \\
        \midrule
        \mt{StopScheduledArrival} & \dtu{Continuous. The scheduled arrival time of the bus at the stop, expressed as seconds since midnight.}{Z-score normalised on the train split, then passed through a small MLP to produce a time encoding vector. The time encoding is shared between the encoder and decoder.} \\
        \midrule
        \mt{Latitude}, \mt{Longitude} & \dtu{Continuous. The WGS-84 latitude and longitude coordinates of the stop location in decimal degrees.}{Combined and projected to a metric CRS, then used to compute pairwise geographic distance matrices between stops. These matrices serve as relational node features for the graph autoencoder, which produces the precomputed stop graph embeddings.} \\
        \midrule
        \mt{StopType} & \dtu{Categorical. The binary transfer category of the stop, distinguishing between "Transfer stops" (where passengers can transfer to other lines) and "Ordinary stops" (where no transfers are available).}{(1) Mapped to an integer code ($0=$~Ordinary, $1=$~Transfer) and passed as a raw scalar into the stop attributes vector. (2) Used to build the attribute-similarity matrix for the graph autoencoder.} \\
        \midrule
        \mt{TransferStop} & \dtu{Continuous. A count indicating the degree of transfer importance of the stop, reflecting how many alternate lines are accessible from that stop. A value of zero denotes a non-transfer stop.}{(1) Z-score normalised and included in the stop attributes vector. (2) Used to build the attribute-similarity matrix for the graph autoencoder.} \\
        \midrule
        \mt{Line} & \dtu{Categorical. The line or route number of the bus, identifying the service pattern operated during the trip.}{Mapped to an integer code and embedded as part of the trip context. The trip context is broadcast globally across all encoder positions.} \\
        \midrule
        \mt{FLAG_TripDirection} & \dtu{Categorical. A binary flag indicating the direction of travel: 1 denotes travel away from the assigned home stop, and 0 denotes travel towards it.}{Mapped to an integer code and embedded as part of the trip context.} \\
        \midrule
        \mt{nStops} & \dtu{Continuous. The total number of stops on the trip. This also serves as a proxy for route length.}{Z-score normalised and included in the trip context as a continuous scalar.} \\
        \midrule
        \mt{BusType} & \dtu{Categorical. The vehicle type used for the trip, reflecting capacity and operational differences across the fleet.}{Mapped to an integer code and embedded as part of the trip context.} \\
        \midrule \midrule
        & \fcat{Temporal Metadata (4)} \\[3pt]
        \mt{Hour} & \dtu{Cyclic. The hour of the day at which the trip departs, ranging from 0 to 23.}{Encoded as sine and cosine components to preserve the cyclic nature of the 24-hour clock. The resulting two-dimensional encoding is included in the trip context.} \\
        \midrule
        \mt{DayType} & \dtu{Cyclic. The day of the week, capturing intra-week variation in travel demand patterns.}{Encoded as sine and cosine components to preserve the cyclic weekly pattern. The resulting two-dimensional encoding is included in the trip context.} \\
        \midrule
        \mt{FLAG_Workday} & \dtu{Categorical. A binary flag indicating whether the day is a workday, distinguishing regular working days from weekends and public holidays.}{Mapped to an integer code and embedded as part of the trip context.} \\
        \bottomrule
    \end{tabular*}
\end{table}

\begin{table}[pos=H]
    \centering \renewcommand{\arraystretch}{1.25} \footnotesize
    \ContinuedFloat \caption{\emph{Cont.}}
    \begin{tabular*}{\textwidth}{p{0.17\textwidth}p{0.78\textwidth}}
        \toprule
        \bt{Feature} & \bt{Description, Transformation, and Usage} \\
        \midrule
        \mt{FLAG_Holiday} & \dtu{Categorical. A binary flag indicating whether the day falls on a public holiday or an observed rest day.}{Mapped to an integer code and embedded as part of the trip context.} \\
        \midrule \midrule
        & \fcat{Urban Characteristics (10)} \\[3pt]
        \mt{luPopulation} & \dtu{Continuous. The resident population count in the grid cells within a 250 m radius of the stop, derived from SSB population statistics at 250 m resolution.}{(1) Z-score normalised and included in the stop attributes vector. (2) Used to build the attribute-similarity matrix for the graph autoencoder.} \\
        \midrule
        \mt{luDwelling} & \dtu{Continuous. The total number of residential dwellings in the grid cells within a 250 m radius of the stop, derived from SSB housing statistics at 250 m resolution.}{(1) Z-score normalised and included in the stop attributes vector. (2) Used to build the attribute-similarity matrix for the graph autoencoder.} \\
        \midrule
        \mt{luBusiness} & \dtu{Continuous. The total number of registered businesses in the grid cells within a 250 m radius of the stop, derived from Statistics Norway (SSB) business statistics at 250 m resolution.}{(1) Z-score normalised and included in the stop attributes vector. (2) Used to build the attribute-similarity matrix for the graph autoencoder.} \\
        \midrule
        \mt{luEmployee} & \dtu{Continuous. The total number of employees across all businesses in the grid cells within a 250 m radius of the stop, derived from SSB business statistics at 250 m resolution.}{(1) Z-score normalised and included in the stop attributes vector. (2) Used to build the attribute-similarity matrix for the graph autoencoder.} \\
        \midrule
        \mt{luBuiAll} & \dtu{Continuous. The total built floor area across all building classifications in the grid cells within a 250 m radius of the stop, derived from SSB building area statistics at 250 m resolution.}{(1) Z-score normalised and included in the stop attributes vector. (2) Used to build the attribute-similarity matrix for the graph autoencoder.} \\
        \midrule
        \mt{luBuiDwell} & \dtu{Continuous. The built floor area attributed to residential dwellings in the grid cells within a 250 m radius of the stop.}{(1) Z-score normalised and included in the stop attributes vector. (2) Used to build the attribute-similarity matrix for the graph autoencoder.} \\
        \midrule
        \mt{luBuiOffice} & \dtu{Continuous. The built floor area attributed to office and business premises in the grid cells within a 250 m radius of the stop.}{(1) Z-score normalised and included in the stop attributes vector. (2) Used to build the attribute-similarity matrix for the graph autoencoder.} \\
        \midrule
        \mt{luBuiHotel} & \dtu{Continuous. The built floor area attributed to hotel and restaurant premises in the grid cells within a 250 m radius of the stop.}{(1) Z-score normalised and included in the stop attributes vector. (2) Used to build the attribute-similarity matrix for the graph autoencoder.} \\
        \midrule
        \mt{luBuiEduEnt} & \dtu{Continuous. The built floor area attributed to educational, entertainment and religious buildings in the grid cells within a 250 m radius of the stop.}{(1) Z-score normalised and included in the stop attributes vector. (2) Used to build the attribute-similarity matrix for the graph autoencoder.} \\
        \midrule
        \mt{luBuiMedical} & \dtu{Continuous. The built floor area attributed to hospital and institutional care buildings in the grid cells within a 250 m radius of the stop.}{(1) Z-score normalised and included in the stop attributes vector. (2) Used to build the attribute-similarity matrix for the graph autoencoder.} \\
        \midrule \midrule
        & \fcat{Natural Terrain (6)} \\[3pt]
        \mt{StopElevation} & \dtu{Continuous. The elevation of the stop above sea level in metres, derived from a digital elevation model (DEM).}{(1) Z-score normalised and included in the stop attributes vector. (2) Used to build the attribute-similarity matrix for the graph autoencoder.} \\
        \bottomrule
    \end{tabular*}
\end{table}

\begin{table}[pos=H]
    \centering \renewcommand{\arraystretch}{1.25} \footnotesize
    \ContinuedFloat \caption{\emph{Cont.}}
    \begin{tabular*}{\textwidth}{p{0.17\textwidth}p{0.78\textwidth}}
        \toprule
        \bt{Feature} & \bt{Description, Transformation, and Usage} \\
        \midrule
        \mt{StopDistance} & \dtu{Continuous. The distance in metres between the current stop and the following stop along the route.}{(1) Z-score normalised and included in the stop attributes vector. (2) Used as a cosine similarity node feature for the graph autoencoder.} \\
        \midrule
        \mt{diffElevation}, \mt{absElevation} & \dtu{Continuous. The signed and absolute elevation change in metres between the current stop and the following stop, capturing the direction and magnitude of the road gradient respectively.}{(1) Z-score normalised and included in the stop attributes vector. (2) Used to build the attribute-similarity matrix for the graph autoencoder.} \\
        \midrule
        \mt{typeElevation} & \dtu{Categorical. The terrain type of the route segment from the current stop to the following stop, encoded as an ordinal: $-1=$~Downhill, $0=$~Flat, $1=$~Uphill.}{(1) Averaged across the stop's observed route segments and included as a numeric scalar in the stop attributes vector. (2) Used to build the attribute-similarity matrix for the graph autoencoder.} \\
        \midrule
        \mt{slopeElevation} & \dtu{Continuous. The gradient of the route segment from the current stop to the following stop, expressed as a percentage and computed as the ratio of the elevation change to the horizontal distance between consecutive stops.}{(1) Z-score normalised and included in the stop attributes vector. (2) Used to build the attribute-similarity matrix for the graph autoencoder.} \\
        \midrule \midrule
        & \fcat{Weather Conditions (5)} \\[3pt]
        \mt{dwTempMin}, \mt{dwTempMax} & \dtu{Continuous. The minimum and maximum recorded temperatures in degrees Celsius for the day.}{Z-score normalised and included in the trip context as continuous scalars.} \\
        \midrule
        \mt{dwTempFeels} & \dtu{Continuous. The apparent temperature in degrees Celsius for the day, accounting for wind chill and humidity effects on perceived warmth.}{Z-score normalised and included in the trip context as a continuous scalar.} \\
        \midrule
        \mt{dwConditions} & \dtu{Categorical. A short textual summary of the prevailing weather conditions for the day.}{Mapped to an integer code and embedded as part of the trip context.} \\
        \midrule
        \mt{dwWindGust} & \dtu{Continuous. The maximum instantaneous wind gust speed in km/h for the day.}{Z-score normalised and included in the trip context as a continuous scalar.} \\
        \midrule \midrule
        & \fcat{Target Variables (4)} \\[3pt]
        \mt{PC_Boarding}, \mt{PC_Alighting} & \dtu{Continuous. The number of passengers boarding and alighting the bus at the stop, recorded by the automated passenger counting system. These are the primary prediction targets for the decoder.}{Transformed via \mt{log1p} followed by z-score normalisation to reduce skewness. (1) Historical stop-level demand series form the inter-stop correlation matrix used in graph embeddings. (2) Per-trip sequences at time step $t-1$ serve as decoder inputs during teacher-forced training and autoregressive inference; the sequence at time step $t$ is the primary decoder output.} \\
        \midrule
        \mt{ST_Delay} & \dtu{Continuous. The arrival delay of the bus at the stop in seconds, defined as the difference between the actual and scheduled arrival times. This is a surrogate prediction target for the encoder.}{Clipped at zero, then transformed via \mt{log1p} followed by z-score normalisation. Predicted by the encoder as a surrogate task to encourage learning of operationally relevant stop representations.} \\
        \midrule
        \mt{ST_Dwell} & \dtu{Continuous. The dwell time of the bus at the stop in seconds. This is a surrogate prediction target for the encoder.}{Transformed via \mt{log1p} followed by z-score normalisation. Predicted by the encoder as a surrogate task to encourage learning of operationally relevant stop representations.} \\
        \bottomrule
    \end{tabular*}
\end{table}

\section{Model Configurations}
\label{app:model-configs}
This section reports the architecture and training settings used for \mt{SMT-GraphFormer} and the stop-level benchmark models. \autoref{tab:config-smt} gives the \mt{SMT-GraphFormer} configuration, \autoref{tab:config-xgb} lists the XGBoost hyperparameters, and \autoref{tab:config-rtdl} summarises the RTDL model variants. The RTDL models use the same training setup: AdamW optimisation with learning rate $3 \times 10^{-4}$, weight decay $10^{-5}$, batch size 256, a maximum of 100 epochs, and early stopping patience of 20 epochs.

\begin{table}[pos=H]
    \centering \renewcommand{\arraystretch}{1.25} \footnotesize
    \caption{Model architecture and training configuration for the SMT-GraphFormer model. \it{Note: All optional components (graph embeddings, stop features, context encoding, surrogate tasks, and MMoE) are enabled by default.}}
    \label{tab:config-smt}
    \begin{tabular*}{0.60\textwidth}{l@{\extracolsep{\fill}}r}
        \toprule
        \bt{Parameter} & \bt{Value} \\
        \midrule
        Embedding Dimension ($d_\text{embed}$) & 256 \\
        Number of Attention Heads ($N_\text{head}$) & 8 \\
        Number of Transformer Layers ($N_\text{layer}$) & 4 \\
        Number of MMoE Experts ($N_\text{expert}$) & 3 \\
        Dropout (Hidden) & 0.10 \\
        \midrule
        Optimiser & AdamW \\
        Learning Rate & $3 \times 10^{-4}$ \\
        Weight Decay & $10^{-2}$ \\
        Maximum Gradient Norm & 1.0 \\
        LR Warmup Fraction & 0.075 \\
        LR Schedule (Post-Warmup) & Cosine Annealing \\
        \midrule
        Batch Size & 256 \\
        Maximum Epochs & 40 \\
        Early Stopping Patience & 20 \\
        Surrogate Loss Weight ($w_\text{surrogate}$) & 1.0 \\
        \midrule
        Graph Autoencoder: Training Epochs & 4,096 \\
        Graph Autoencoder: Optimiser & Adam \\
        Context Encoder: Embedding Dimension ($d_\text{context}$) & 16 \\
        Scheduled Sampling: Maximum Probability ($p_\text{max}$) & 0.5 \\
        \bottomrule
    \end{tabular*}
\end{table}

\begin{table}[pos=H]
    \centering \renewcommand{\arraystretch}{1.25} \footnotesize
    \caption{Training configuration for the XGBoost benchmark model. \it{Note: The meta-regressor trains a dedicated set of trees per target within a single coordinated boosting process.}}
    \label{tab:config-xgb}
    \begin{tabular*}{0.60\textwidth}{l@{\extracolsep{\fill}}r}
        \toprule
        \bt{Parameter} & \bt{Value} \\
        \midrule
        Objective & \mt{reg:squarederror} \\
        Multi-Output Strategy & \mt{one_output_per_tree} \\
        Tree Method & \mt{hist} \\
        Categorical Features & \mt{true} \\
        \midrule
        Number of Estimators & 2048 \\
        Early Stopping Rounds & 205 \\
        Learning Rate ($\eta$) & 0.04 \\
        Maximum Tree Depth & 16 \\
        Minimum Child Weight & 8.5 \\
        Minimum Split Loss ($\gamma$) & 0.12 \\
        \midrule
        Row Subsampling & 0.90 \\
        Column Subsampling per Tree & 0.50 \\
        L1 Regularisation ($\alpha$) & 0.001 \\
        L2 Regularisation ($\lambda$) & 0.0 \\
        \bottomrule
    \end{tabular*}
\end{table}

\begin{table}[pos=H]
    \centering \renewcommand{\arraystretch}{1.25} \footnotesize
    \caption{Model architecture configurations for the RTDL benchmark models: MLP, ResNet, and FT-Transformer.}
    \label{tab:config-rtdl}
    \begin{threeparttable}
        \begin{tabular*}{0.85\textwidth}{l@{\extracolsep{\fill}}rrr}
            \toprule
            \bt{Parameter} & \bt{MLP} & \bt{ResNet} & \bt{FT-Transformer} \\
            \midrule
            Number of Blocks / Layers & 2 & 2 & 3 \\
            Block Width ($d_\text{block}$) & 384 & 192 & 192 \\
            Hidden Multiplier & -- & 2.0 & 1.333\tnote{1} \\
            Activation & ReLU & ReLU & ReGLU \\
            Number of Attention Heads & -- & -- & 8 \\
            \midrule
            Dropout (Hidden) & 0.10 & 0.15 & 0.10\tnote{2} \\
            Dropout (Residual) & -- & 0.0 & 0.0 \\
            Dropout (Attention) & -- & -- & 0.20 \\
            \midrule
            Output Dimension & 4 & 4 & 4 \\
            Categorical Handling & One-Hot Encoding & One-Hot Encoding & Learned Embeddings \\
            \bottomrule
        \end{tabular*}
        \begin{tablenotes}
            \item[1] Effective hidden dimension is 256 as the first linear layer outputs 512 which ReGLU halves.
            \item[2] This dropout is applied to feedforward layers only, separate from the attention dropout.
        \end{tablenotes}
    \end{threeparttable}
\end{table}

\section{Detailed Predictive Performance}
\label{app:benchmark-details}
This section reports RMSE, MAE, and $R^2$ across the training, validation, and test splits. \autoref{tab:results-xgb} gives the XGBoost benchmark results, \autoref{tab:results-rtdl} gives the RTDL benchmark results for MLP, ResNet, and FT-Transformer, and \autoref{tab:results-smt} gives the \mt{SMT-GraphFormer} results under teacher-forced and autoregressive inference. Delay and dwell metrics are identical across the two \mt{SMT-GraphFormer} inference protocols because these outputs are produced by the encoder.

\begin{table}[pos=H]
    \centering \renewcommand{\arraystretch}{1.25} \footnotesize
    \caption{Predictive performance of the XGBoost benchmark model across data splits.}
    \label{tab:results-xgb}
    \begin{tabular*}{0.55\textwidth}{l@{\extracolsep{\fill}}l|ccc}
        \toprule
        \bt{Target} & \bt{Data Split} & \bt{RMSE} & \bt{MAE} & \bm{R^2} \\
        \midrule
        Boarding & Training & 2.081 & 0.927 & 0.581 \\
        & Validation & 2.180 & 1.048 & 0.459 \\
        & Test & 1.987 & 0.954 & 0.487 \\
        \midrule
        Alighting & Training & 2.211 & 0.932 & 0.552 \\
        & Validation & 2.288 & 1.074 & 0.417 \\
        & Test & 2.092 & 0.972 & 0.449 \\
        \midrule
        Delay & Training & 128.25 & 66.77 & 0.333 \\
        & Validation & 131.21 & 75.42 & 0.125 \\
        & Test & 121.44 & 73.57 & 0.115 \\
        \midrule
        Dwell & Training & 32.70 & 8.40 & 0.411 \\
        & Validation & 35.37 & 9.36 & 0.300 \\
        & Test & 34.62 & 9.12 & 0.311 \\
        \bottomrule
    \end{tabular*}
\end{table}

\newcommand{\mcbt}[1]{\multicolumn{3}{c}{\bt{#1}}}
\begin{table}[pos=H]
    \centering \renewcommand{\arraystretch}{1.25} \footnotesize
    \caption{Predictive performance of the RTDL benchmark models across data splits.}
    \label{tab:results-rtdl}
    \begin{tabular*}{0.95\textwidth}{l@{\extracolsep{\fill}}l|ccc|ccc|ccc}
        \toprule
        \bt{Target} & \bt{Data Split} & \mcbt{MLP} & \mcbt{ResNet} & \mcbt{FT-Transformer} \\
        & & \bt{RMSE} & \bt{MAE} & \bm{R^2} & \bt{RMSE} & \bt{MAE} & \bm{R^2} & \bt{RMSE} & \bt{MAE} & \bm{R^2} \\
        \midrule
        Boarding & Training & 2.225 & 0.979 & 0.521 & 2.200 & 0.983 & 0.532 & 2.198 & 0.983 & 0.533 \\
        & Validation & 2.115 & 1.008 & 0.490 & 2.093 & 1.000 & 0.501 & 2.072 & 0.993 & 0.511 \\
        & Test & 1.992 & 0.952 & 0.484 & 1.987 & 0.955 & 0.486 & 1.939 & 0.935 & 0.511 \\
        \midrule
        Alighting & Training & 2.398 & 0.998 & 0.474 & 2.349 & 0.996 & 0.495 & 2.359 & 0.992 & 0.491 \\
        & Validation & 2.214 & 1.024 & 0.454 & 2.192 & 1.014 & 0.464 & 2.157 & 1.000 & 0.482 \\
        & Test & 2.110 & 0.964 & 0.440 & 2.111 & 0.964 & 0.439 & 2.040 & 0.940 & 0.476 \\
        \midrule
        Delay & Training & 134.23 & 71.35 & 0.269 & 139.34 & 73.45 & 0.213 & 129.15 & 70.72 & 0.324 \\
        & Validation & 136.70 & 79.14 & 0.050 & 134.62 & 77.43 & 0.079 & 128.62 & 74.71 & 0.159 \\
        & Test & 122.77 & 74.69 & 0.096 & 126.10 & 76.71 & 0.046 & 119.26 & 72.38 & 0.147 \\
        \midrule
        Dwell & Training & 35.19 & 8.98 & 0.318 & 35.38 & 9.09 & 0.311 & 35.67 & 9.08 & 0.299 \\
        & Validation & 35.33 & 9.40 & 0.301 & 35.45 & 9.39 & 0.297 & 35.50 & 9.29 & 0.295 \\
        & Test & 34.67 & 9.24 & 0.309 & 34.87 & 9.29 & 0.301 & 34.90 & 9.15 & 0.300 \\
        \bottomrule
    \end{tabular*}
\end{table}

\begin{table}[pos=H]
    \centering \renewcommand{\arraystretch}{1.25} \footnotesize
    \caption{Split-wise predictive performance of the SMT-GraphFormer model under teacher-forced (TF) and autoregressive (AR) inference. \it{Note: Delay and dwell are produced by the encoder and are identical under both inference protocols.}}
    \label{tab:results-smt}
    \begin{tabular*}{0.75\textwidth}{l@{\extracolsep{\fill}}llccc}
        \toprule
        \bt{Protocol} & \bt{Target} & \bt{Data Split} & \bt{RMSE} & \bt{MAE} & \bm{R^2} \\
        \midrule
        Teacher-forced (TF) & Boarding & Training & 1.901 & 0.887 & 0.651 \\
        & & Validation & 1.978 & 0.940 & 0.554 \\
        & & Test & 1.861 & 0.892 & 0.550 \\
        \cmidrule{2-6}
        & Alighting & Training & 1.538 & 0.732 & 0.783 \\
        & & Validation & 1.597 & 0.786 & 0.716 \\
        & & Test & 1.505 & 0.743 & 0.715 \\
        \midrule
        Autoregressive (AR) & Boarding & Training & 2.082 & 0.945 & 0.581 \\
        & & Validation & 2.063 & 0.979 & 0.515 \\
        & & Test & 1.932 & 0.926 & 0.515 \\
        \cmidrule{2-6}
        & Alighting & Training & 2.269 & 0.970 & 0.529 \\
        & & Validation & 2.170 & 0.997 & 0.475 \\
        & & Test & 2.044 & 0.941 & 0.474 \\
        \midrule
        TF or AR & Delay & Training & 121.59 & 66.58 & 0.400 \\
        & & Validation & 130.93 & 76.03 & 0.129 \\
        & & Test & 116.50 & 70.96 & 0.186 \\
        \cmidrule{2-6}
        & Dwell & Training & 34.21 & 8.91 & 0.355 \\
        & & Validation & 34.43 & 9.24 & 0.337 \\
        & & Test & 33.74 & 9.11 & 0.346 \\
        \bottomrule
    \end{tabular*}
\end{table}

\end{document}